\documentclass{article}
\PassOptionsToPackage{numbers,compress}{natbib}

 \usepackage[main, final]{neurips_2026}

\usepackage[utf8]{inputenc} 
\usepackage[T1]{fontenc}    
\usepackage{hyperref}       
\usepackage{url}            
\usepackage{booktabs}       
\usepackage{amsfonts}       
\usepackage{nicefrac}       
\usepackage{microtype}      
\usepackage{xcolor}         
\usepackage{xcolor}
\usepackage[table]{xcolor} 
\usepackage{colortbl}
\usepackage{graphicx}      
\usepackage{multirow}
\usepackage{caption}
\usepackage{wrapfig}
\usepackage{makecell}
\usepackage[most]{tcolorbox}
\usepackage{xcolor}
\usepackage{alltt}
\usepackage[hybrid]{markdown}
\usepackage{enumitem}
\usepackage{listings}
\usepackage{upquote}
\usepackage{needspace}
\usepackage{placeins}

\title{Personal Visual Memory from Explicit and Implicit Evidence}

\definecolor{thaoblue}{HTML}{0F52BA}

\author{Viet Nguyen$^{1}$ \quad \quad Thao Nguyen$^{2}$ \quad \quad Vishal M. Patel$^{1}$\thanks{Equal advising.} \quad \quad Yuheng Li$^{3}$\footnotemark[1] \\ \\
$^{1}$Johns Hopkins University \quad 
$^{2}$University of Wisconsin-Madison \quad
$^{3}$Adobe Research\\ \\
\url{https://viettmab.github.io/visualmem-page/}
}

\begin{document}
\maketitle
\begin{figure}[h]
    \centering
    \includegraphics[width=0.95\linewidth]{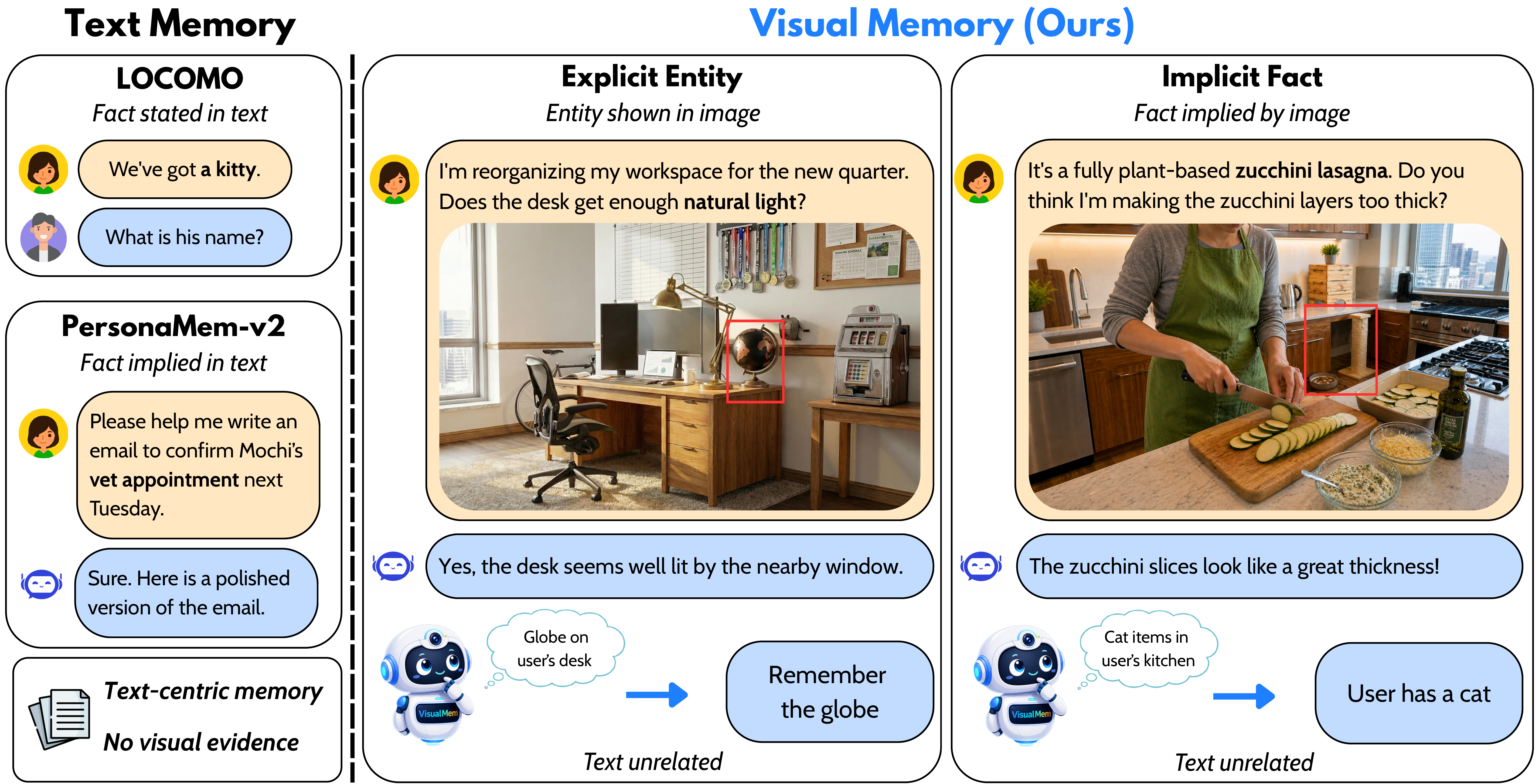}
    \caption{\textbf{Comparison between prior text-memory benchmarks and our visual-memory setting.} Existing benchmarks mainly test facts that are explicit in text or implied by text alone. In contrast, \texttt{VisualMem} introduces two visually grounded memory challenges: \emph{explicit entity}, where the model must remember a recurring visual entity and its identity from prior images, and \emph{implicit fact}, where the target fact must be inferred from visual cues while the conversation is unrelated to it.}
    \label{fig:teaser}
\end{figure}
\begin{abstract}
Long-term memory is increasingly important for personalized AI agents, yet existing benchmarks and methods remain largely text-centric. Even when images are included, the user-specific information needed for later questions is typically recoverable from text alone, and most memory systems reduce image turns to generic captions. Yet images often carry personal information that text rarely states---both \emph{explicit} evidence, such as recurring user-associated entities, and \emph{implicit} evidence, such as latent user facts inferred from visual or multimodal cues. We introduce a benchmark for \emph{personal visual memory} that targets both forms of evidence, and propose \textsc{VisualMem}, a hybrid visual--text architecture that augments a text-memory backend with a structured personal visual memory module. Rather than collapsing images into captions, \textsc{VisualMem}  uses conversational context to resolve identity, ownership, and durable user facts. Experiments show that \textsc{VisualMem} substantially outperforms prior memory systems on our benchmark while remaining competitive on standard text-memory benchmarks, indicating that personal visual memory is a distinct and important component of long-term memory for personalized AI agents.
\end{abstract}
\section{Introduction}
Human experience is inherently multimodal: the personal details that define our lives---our homes, relationships, belongings, and routines---are expressed not only through language, but also through the images we capture and share. As AI assistants become increasingly integrated into everyday life, they are exposed to growing amounts of personal multimodal data across extended interactions. A truly personalized assistant must therefore be able to remember and reason over both text and images. Visual memory can matter in two complementary ways, as illustrated in Fig.~\ref{fig:teaser}. In the explicit case (middle), users may share photos containing recurring entities, such as a personal object on their desk, and expect the assistant to recognize them later. In the implicit case (right), personal facts must be inferred from visual cues: for example, a cat bowl and scratching post in a kitchen may suggest that the user owns a cat, even if the conversation is not about pets. Such cases require agents to retain visually grounded personal information beyond what is explicitly stated in text.

Recent work has advanced long dialogue reasoning, memory evaluation, and memory-augmented agents~\cite{locomo,longmemeval,personamem2,mem0,memos}. However, current long-term memory research remains largely \emph{text-centric}. Existing benchmarks may include images, but the information needed to answer later questions is often recoverable from text alone~\cite{locomo,longmemeval,personamem2}. This limitation is also reflected in the design of existing memory systems. Most approaches handle image turns by converting them into generic captions and then processing them with a text-only memory pipeline~\cite{mem0,memos}. This approach has two key limitations. First, at the \emph{single-image level}, captions are lossy: they may omit identity, visual details, and ownership cues that are crucial for personalized visual knowledge. Second, at the \emph{cross-image level}, isolated captions make it difficult to reason across images scattered over long conversations. For example, a single photo may not reveal whether a cat  belongs to the user, but repeated visual evidence over time can resolve such ambiguity.

A key challenge in studying visual memory is obtaining suitable evaluation data. Such data must resemble natural conversations with long histories, recurring entities, consistent private environments, delayed questions, and reliable visual grounding. Collecting it from real users would require sustained access to personal multimodal histories and careful annotation of visually grounded private facts, making the process costly, difficult to scale, and privacy-sensitive~\cite{lee2024towards,tomekcce2024private}. Recent progress in controllable and personalized image generation offers a practical alternative: synthetic images can preserve subject-level consistency across varied contexts and can support both evaluation and downstream representation learning~\cite{he2024imagineyourself,sundaram2024personalized,peng2025dreambenchpp}. We therefore adopt a synthetic construction pipeline that enables controlled generation of explicit entities, implicit facts, distractors, cross-image consistency, and grounded questions at scale. Based on this pipeline, we introduce a benchmark where answering test questions requires retaining and reasoning over visual information from earlier interactions, rather than relying on text alone. Our benchmark covers two settings: an \emph{explicit} setting, where agents must remember recurring visually grounded entities, and an \emph{implicit} setting, where agents must infer latent personal facts from visual or multimodal evidence.

Beyond evaluation, these data requirements reveal a broader systems challenge: visual information should not be reduced to generic captions, but should instead be stored as persistent evidence that can be revisited and refined as new context arrives. To address this challenge, we propose \textsc{VisualMem}, a personal visual--text memory framework designed to integrate with existing text-based memory systems~\cite{mem0,memos}. For textual inputs, \textsc{VisualMem} preserves the original text memory pipeline. For image inputs, it uses a dedicated visual memory pipeline with a two-stage contextual decision process. When an image is first observed, the system considers both the image and its surrounding session context to determine whether its personal relevance is clear. If so, it extracts and stores structured personal information in both textual and visual formats; otherwise, the image is placed into a \emph{pending} state. As more visual memories accumulate, \textsc{VisualMem} revisits pending images and compares them against prior observations, enabling it to progressively disambiguate identity, ownership, and other personal facts. Our experiments validate the importance of explicitly modeling visual memory. Caption-only methods perform substantially worse on our benchmark, where questions require reasoning over prior visual evidence, while \textsc{VisualMem} more effectively captures and aggregates visual information over time. At the same time, \textsc{VisualMem} remains competitive on standard text-centric memory benchmarks, demonstrating compatibility with existing memory systems. These results suggest that future memory agents should move beyond treating images as auxiliary text and instead explicitly model the persistent personal information contributed by visual interactions.

Our contributions are summarized as follows:
\begin{itemize}
    \item We identify a limitation of current long-term memory research: existing benchmarks and methods remain primarily text-centric, even in multimodal settings.
    \item We introduce a benchmark for visual memory in AI agents in which later questions depend on retaining visually grounded personal information.
    \item We propose \textsc{VisualMem}, a hybrid visual--text memory architecture that stores structured personal visual memory beyond generic captions.
    \item We show that \textsc{VisualMem} performs best on the new visual benchmark while remaining competitive on standard text-memory benchmarks.
\end{itemize}
\section{Related Work}
\textbf{Memory Benchmarks.}
Recent work has introduced benchmarks for evaluating long-term memory in conversational agents and personalized assistants, covering long dialogue histories, personal facts, multi-session recall, temporal reasoning, knowledge updates, abstention, and implicit personalization~\cite{locomo,longmemeval,personamem2,memorybank,perltqa}. These benchmarks primarily evaluate whether agents can retain and reason over textual information across long horizons. Although some benchmarks include multimodal interactions~\cite{locomo}, images are typically used as auxiliary context rather than as the primary source of decisive evidence. A complementary line of work studies long-context visual retrieval and multi-image reasoning through needle-in-a-haystack-style evaluations~\cite{wang2024needle,wang2025multimodal,wu2024visual}. These benchmarks show that MLLMs still struggle to retrieve relevant visual evidence from long multimodal contexts and to reason across many images. However, they focus mainly on visual context retrieval or multi-image question answering, rather than personal memory. In contrast, our benchmark targets visually grounded personal memory, where decisive evidence is primarily visual and questions require either explicit recurring-entity grounding or latent multimodal inference.

\textbf{RAG- and Memory-Based Methods.}
A broad line of work augments LLMs with external memory through retrieval-augmented generation (RAG)~\cite{shuster2021retrieval,ram2023incontext,shi2024replug} or explicit memory modules~\cite{mem0,memos,memorybank,meminsight,memverse,lightmem,simplemem,memgpt,amem}. 
RAG-based methods typically index past interactions as text chunks, summaries, or extracted facts and retrieve relevant information at query time~\cite{longmemeval,shuster2021retrieval,ram2023incontext}, operating over relatively static representations of the original context. Recent variants improve retrieval through adaptive control~\cite{selfrag} or graph-structured associations~\cite{hipporag2}. 
In contrast, memory-based systems maintain persistent stores where information is structured and updated over time. These systems include graph memories~\cite{mem0}, layered memory abstractions~\cite{memos}, hierarchical multimodal memories~\cite{memverse}, and compressed semantic representations~\cite{simplemem}. Despite their differences, most approaches remain primarily text-oriented: image turns are usually converted into captions or summaries before storage and retrieval. Such representations may support coarse scene recall, but they are less suited to preserving visual details and cross-image consistency cues needed for long-term personalized visual memory. Our method addresses this limitation by maintaining structured visual memories over recurring people, user-associated assets, ownership, and durable visually grounded facts, while remaining compatible with standard text-memory backends.

\textbf{Personalization in MLLMs.}
Recent work on MLLM personalization studies how multimodal assistants adapt to user-specific concepts, preferences, and interaction histories~\cite{wu2024personalized}. 
A major line of work focuses on concept-level personalization, where models learn user-defined visual concepts from a few examples and use them for personalized understanding or generation~\cite{myvlm,yollava,yochameleon}. 
Other methods explore retrieval-based or training-free personalization to incorporate user-specific evidence without extensive model updating~\cite{rap,r2p}. 
More recent work further considers memory-based or temporal personalization, where user attributes and preferences may persist or evolve over time~\cite{personavlm,tame}. These efforts are closely related to ours, but largely emphasize adapting to predefined concepts or tracking explicit attributes. Our work focuses on long-term visual memory: discovering and retaining recurring people, user-associated assets, and implicit personal facts from image evidence, while resolving ownership and identity ambiguity over time. 

\section{Benchmark Construction}

We construct the benchmark through a synthetic pipeline that generates long-horizon multimodal interactions grounded in persistent user personas. Our pipeline has four stages. First, we build a structured \emph{multimodal persona context}, including a user profile, recurring social entities, and user-associated assets (Sec.~\ref{subsec:persona_context}). Second, we generate temporally coherent \emph{event sequences} and synthesize multimodal \emph{conversations} under three modes---implicit, explicit, and distractor---which control the memory signals introduced (Sec.~\ref{subsec:conversation_gen}). Third, we generate \emph{images} conditioned on persistent entities and environments to ensure global visual consistency across interactions (Sec.~\ref{subsec:image_gen}). Finally, based on the constructed multimodal histories, we derive benchmark questions that probe delayed memory over previously observed entity identities and latent personal facts.

\subsection{Multimodal Persona Context}
\label{subsec:persona_context}
Before generating conversations, we construct a structured \emph{persona-centric memory context} that defines the persistent world of each user. This context specifies \emph{who} the user is, \emph{which} entities recur across interactions, and \emph{how} events evolve over time. It has three components:

\textbf{User profile.}
To ensure diverse and representative user characteristics, we sample candidate personas from PersonaHub~\cite{personahub}, a large-scale corpus with broad demographic coverage. Each persona provides a concise description (1--3 sentences) of core traits, which we further enrich with structured attributes (e.g., interests, routines, personal details). We also generate a corresponding profile image for multimodal grounding.

\textbf{Persistent entities.}
We define two types of recurring entities. The first is the user's \emph{social graph} (e.g., family, friends, coworkers), which enables consistent person references and identity-based memory tracking. The second is \emph{user-associated assets} (e.g., personal objects, pets), each described with distinctive visual properties and paired with a reference image for consistent visual appearance.

\textbf{Event structure.}
We generate a sequence of temporally ordered events representing the user's ongoing experiences. Each event includes a timestamp, a short summary, and, when applicable, causal links to prior events. Events span everyday domains such as work, health, relationships, and leisure, and are constrained to remain globally consistent with the user profile. Each event also specifies the relevant social entities and assets, enabling evaluation over recurring people and objects.

\subsection{Conversation Generation}
\label{subsec:conversation_gen}
\begin{figure}[t]
    \centering
    \includegraphics[width=1.0\linewidth]{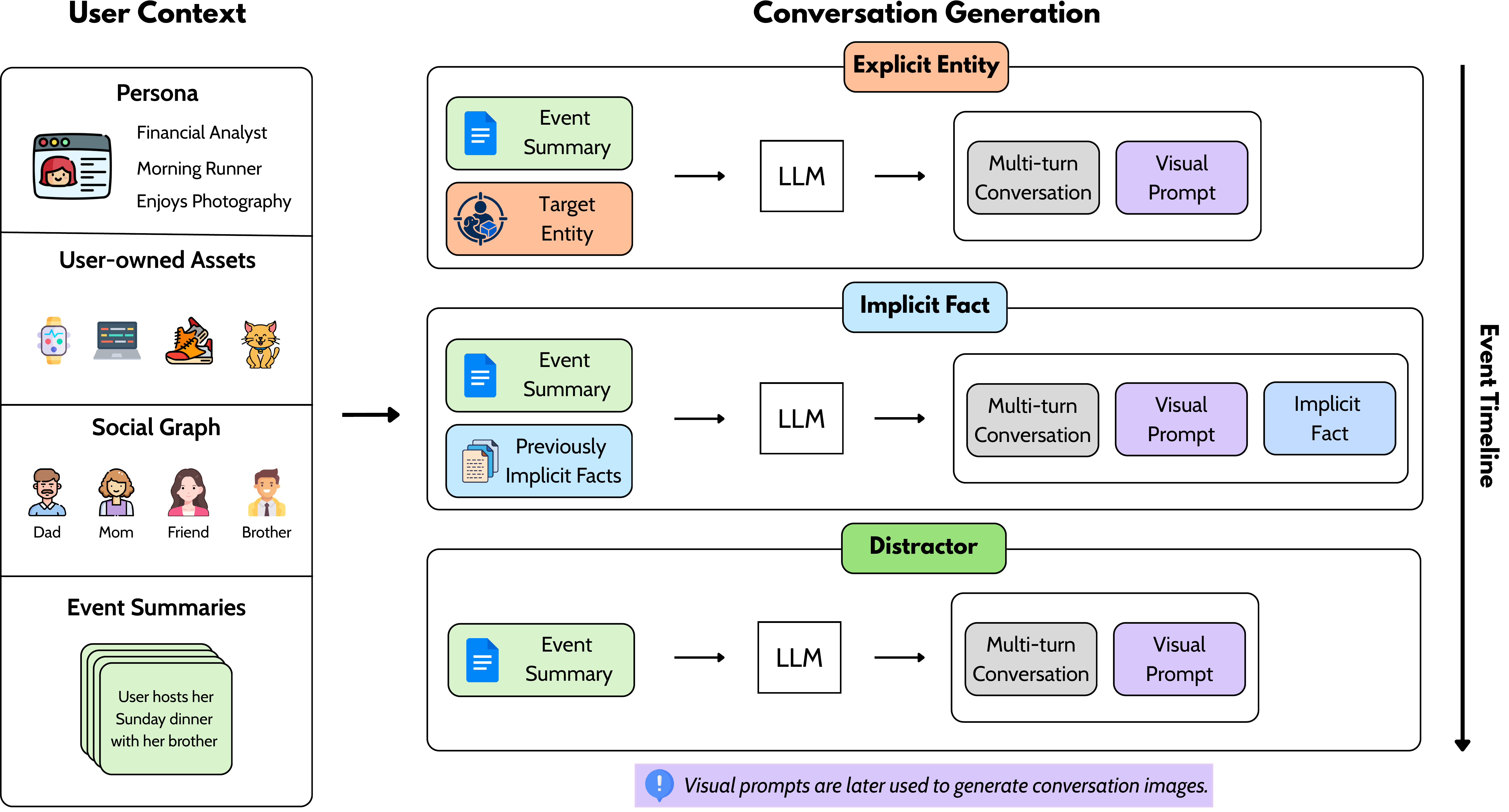}
    \caption{\textbf{Conversation construction from persistent user context.} A persona-centric context, including user profile, recurring social contacts, user-owned assets, and event summaries, is used to generate three families of multimodal conversations: \emph{explicit entity}, \emph{implicit fact}, and \emph{distractor}.}
    \label{fig:conversation_gen}
    \vspace{-5mm}
\end{figure}

Given persona-centric context and event summaries, we generate multimodal conversations that simulate realistic user--assistant interactions. Each session contains a multi-turn dialogue and an associated image, represented by a caption at this stage. Based on the evaluation purpose, there are three conversation types: \emph{explicit entity}, \emph{implicit fact}, and \emph{distractor}. Figure~\ref{fig:conversation_gen} illustrates this process.

\textbf{Explicit entity generation.}
This type of conversation explicitly presents user-associated entities in images (e.g., Fig.~\ref{fig:teaser} Ours, left) to evaluate whether models can remember recurring entities across interactions. Each example is centered on a predefined target entity from the persona context, either a recurring person from the user's social graph (\emph{Target Person}) or a user-associated asset (e.g., object or pet; \emph{Target Asset}). Given the event summary and target entity, an LLM generates both a multi-turn conversation and an image caption that visually includes the target. 

\textbf{Implicit fact generation.}
This mode targets latent personal information that is not explicitly stated in text but can be inferred from visual cues (e.g., Fig.~\ref{fig:teaser} Ours, right). For each event, an LLM generates a multi-turn conversation, an image caption, and a set of implicit facts supported by the interaction. Generation is conditioned on the event summary and previously established implicit facts to maintain global consistency and avoid contradictions over time. We consider two variants. In \emph{visual-only} examples, the decisive evidence appears only in the image, such as a cat tree in the background implying that the user owns a cat. In \emph{multimodal} examples, the inference requires combining image and dialogue signals, such as running gear in an image together with dialogue about post-workout soreness implying a stable exercise habit.

\textbf{Distractor generation.}
This mode introduces non-target interactions for evaluation. We include two types of distractors. \emph{Neutral interactions} consist of conversations and images unrelated to benchmark-critical entities, increasing contextual diversity and sequence length. \emph{Hard negatives} include visually similar but incorrect entities, or scenes resembling user-associated environments that do not belong to the user (e.g., a friend's living room). These examples are designed to test whether models can avoid incorrectly storing irrelevant visual content as user-specific information.

\textbf{Ambiguity of image ownership.}
In realistic interactions, the source or ownership of an image is often ambiguous: a bedroom photo may depict the user's home, someone else's home, or an external reference image. To reflect this, conversations do not always explicitly state image ownership or context. For example, instead of saying ``here is my bedroom,'' the user may simply discuss an unrelated topic while sharing an image. 

To maintain visual consistency, we rely on the generation pipeline (see next) to ensure that user-associated entities and environments remain visually coherent across events, while unrelated images are distinct. As a result, implicit facts and entity-level memory should only be inferred when supported by consistent visual evidence, preventing spurious memory formation.

\subsection{Image Generation and Scene Consistency}
\label{subsec:image_gen}
\begin{figure}[t]
    \centering
    \includegraphics[width=1.0\linewidth]{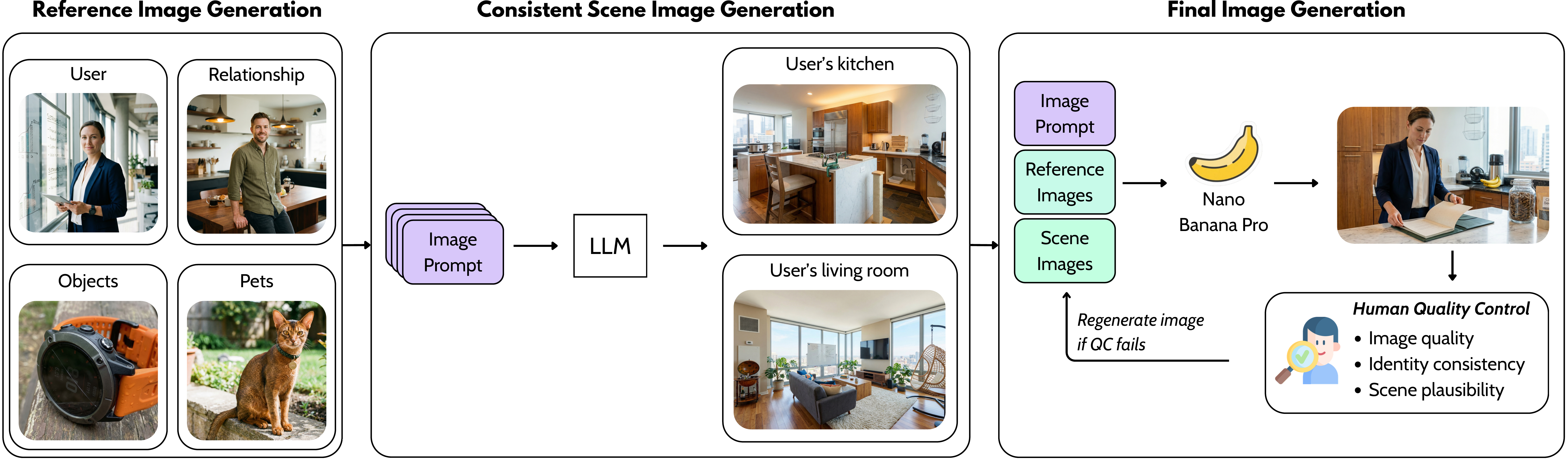}
    \caption{\textbf{Image generation with global consistency.} We first generate reference images for recurring entities, then derive location-level scene references, and finally synthesize conversation images conditioned on both sources, followed by human quality control.}
    \label{fig:image_gen}
    \vspace{-5mm}
\end{figure}
We synthesize images for each conversation from the generated visual prompts, while enforcing global visual consistency at both the entity and scene levels:

\textbf{Entity Consistency via Conditioning.}
To preserve stable visual identity across interactions, we use the reference images of persistent entities from the persona context, including the user, social contacts, and user-associated assets. Whenever a visual prompt involves a specific entity, image synthesis is conditioned on the corresponding reference image.

\textbf{Scene Consistency via Location Grouping.} 
To maintain coherent environments, we group events by semantic location, such as the user's bedroom, the user's office, or a friend's living room. For each location group, an LLM derives a shared scene description capturing stable properties such as layout, lighting, and dominant structures, while excluding transient details. We then generate a reference image for each location and use it as a conditioning signal during image synthesis.

Finally, all generated images are manually reviewed for visual quality, identity fidelity, and consistency with the intended scene semantics. Samples with noticeable artifacts or inconsistencies are filtered out. Figure~\ref{fig:image_gen} summarizes the image generation pipeline, and Appendix~\ref{app:benchmark_validation} describes the complete benchmark validation procedure.


\subsection{Benchmark Statistics}
\begin{figure}[t]
\centering
\begin{minipage}[t]{0.50\textwidth}
    \vspace{0pt}
    \centering
    \includegraphics[width=0.9\textwidth]{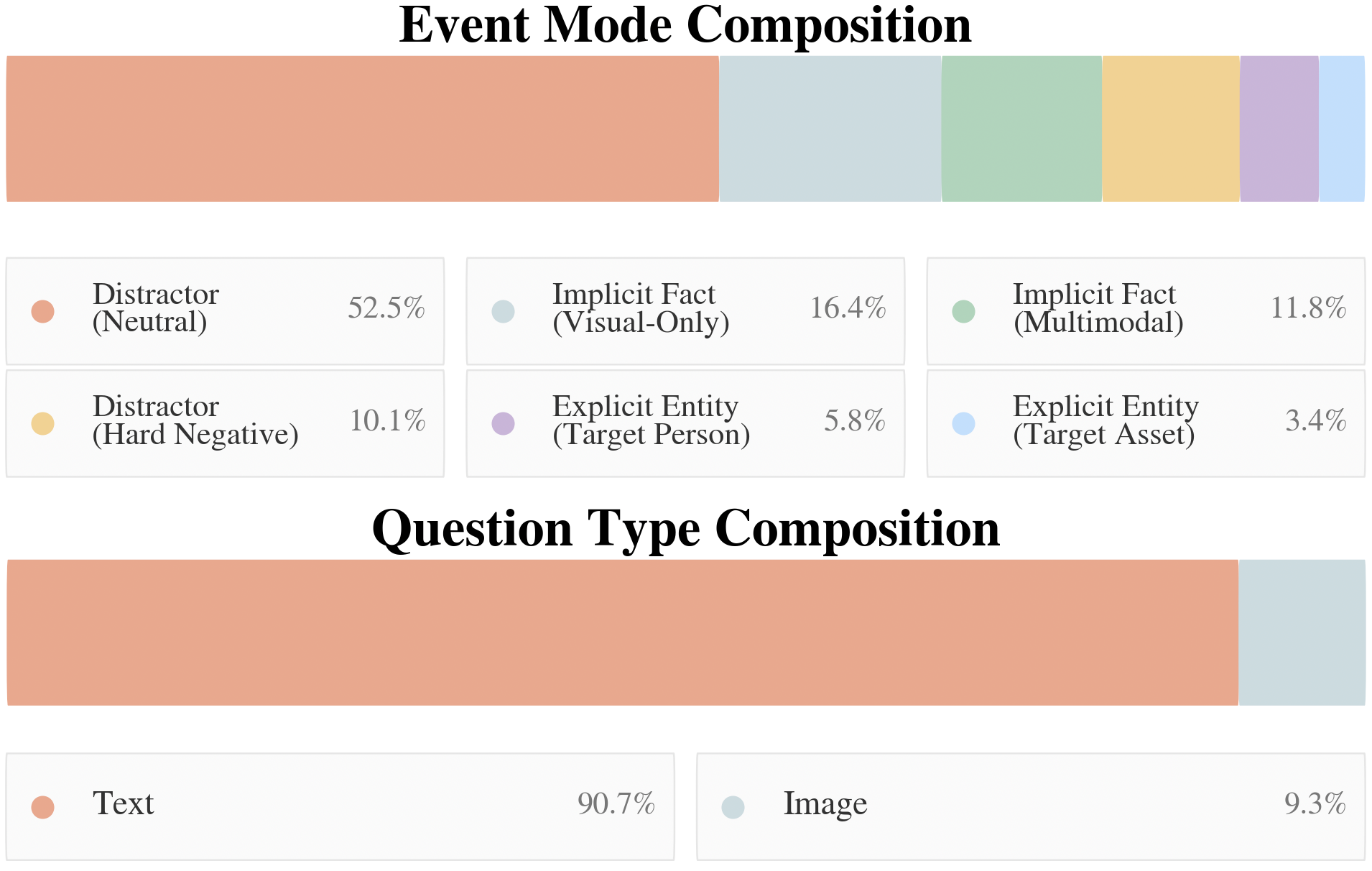}
    \caption{Benchmark composition by event mode and question type.}
    \label{fig:benchmark_composition}
\end{minipage}
\hfill
\begin{minipage}[t]{0.48\textwidth}
    \vspace{0pt}
    \centering
    \small
    \captionof{table}{Benchmark summary statistics.}
    \label{tab:benchmark_summary}
    \begin{tabular}{lr}
    \hline
    \rowcolor{gray!20} Statistics & \# Counts \\
    \hline
    Total \# personas & 10 \\
    Total \# events & 1,717 \\
    Total \# questions & 696 \\
    Total \# images & 1,718 \\
    \hline
    Avg. \# events per persona & 171.7 \\
    Avg. \# topics per persona & 25.0 \\
    Avg. \# assets per persona & 5.8 \\
    Avg. \# social links per persona & 6.1 \\
    Avg. \# questions per persona & 69.6 \\
    Avg. \# turns per event & 11.84 \\
    \hline
    Avg. \# text tokens per persona & 131,110.6 \\
    \hline
    \end{tabular}
\end{minipage}
\end{figure}

Based on the constructed multimodal interaction histories, we generate evaluation questions targeting either recurring entity identity or latent personal facts inferred from prior context in a multiple-choice format.
Our benchmark contains 10 personas, 1,717 events, 1,718 images, and 696 questions. On average, each persona has about 172 events, 25 topics, 6 user-associated assets, and 6 social links. Each event contains 10--12 conversation turns, and the full multimodal history for one persona contains roughly 131K text tokens. Figure~\ref{fig:benchmark_composition} shows the distribution of event modes and question types. Most events are distractors, preserving realistic long-horizon interaction structure, while benchmark-targeted examples cover both explicit-entity and implicit-fact settings. Questions are primarily text-based, with a smaller multimodal subset that additionally includes images. Table~\ref{tab:benchmark_summary} provides the full dataset breakdown.

\section{VisualMem: A Hybrid Visual--Text Memory Framework}

\begin{figure}[t]
    \centering
    \includegraphics[width=1.0\linewidth]{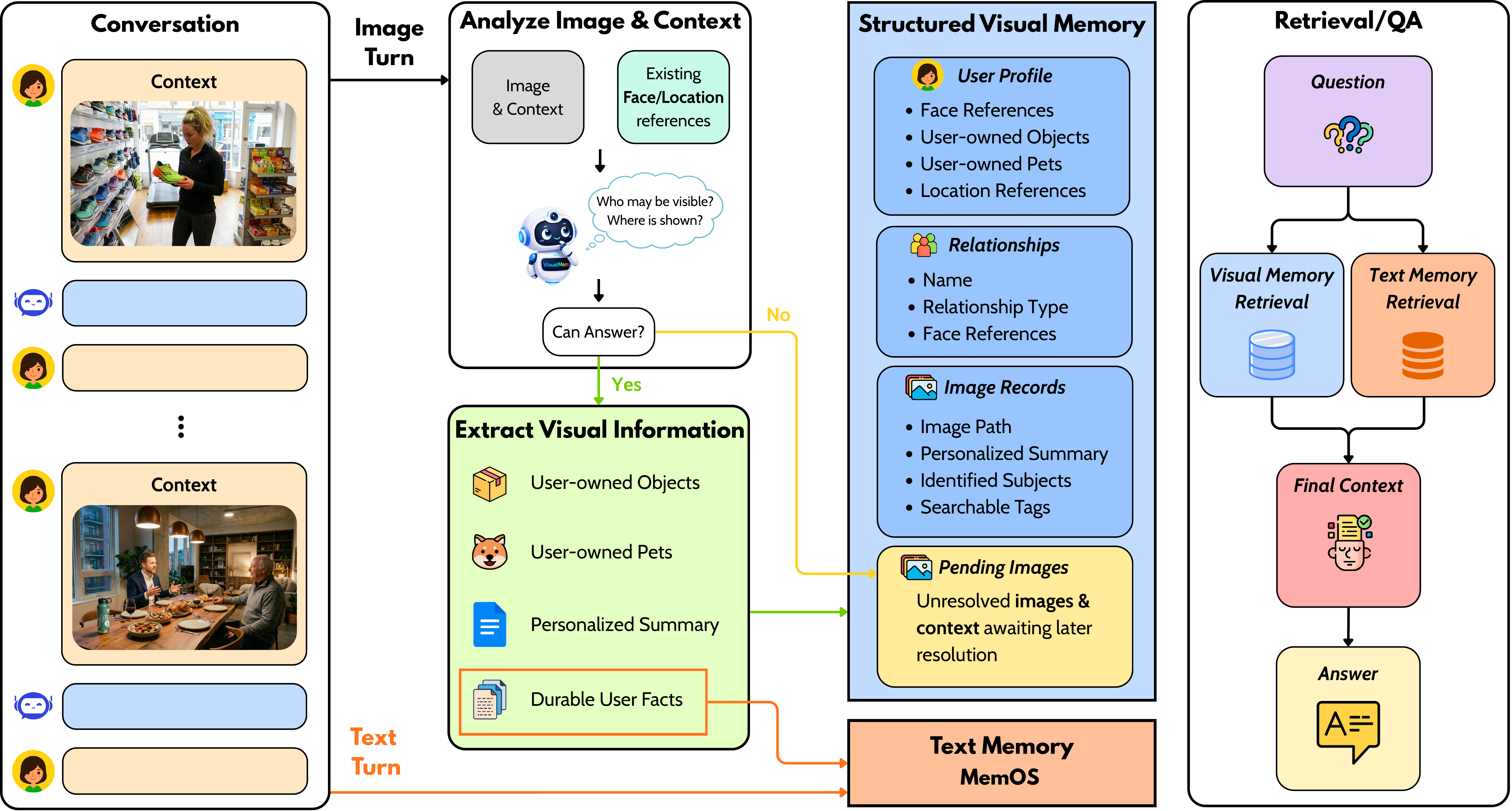}
    \caption{\textbf{Overview of \textbf{\textsc{VisualMem}}.} Given a conversation with image turns, \textsc{VisualMem} jointly analyzes the image and its surrounding context, extracts personalized visual information, stores it in a structured visual memory, and combines visual retrieval with text-memory retrieval for downstream question answering.}
    \label{fig:approach}
    \vspace{-5mm}
\end{figure}

Existing memory systems typically reduce images to captions. As shown in our experiments, this naive strategy performs poorly on multimodal visual memory tasks in our benchmark. To address this limitation, we propose \textsc{VisualMem}, a hybrid visual--text memory framework for long-horizon multimodal interactions (see Fig.~\ref{fig:approach}). \textsc{VisualMem} augments a text-based memory backend with a dedicated visual memory module, forming a plug-and-play framework that can be seamlessly integrated with existing memory systems~\cite{mem0, memos}.


\subsection{Building Visual Memory}

\textsc{VisualMem} processes multimodal interactions in a modality-aware manner. Text-only turns are handled directly by the text-memory backend, while image turns trigger a dedicated visual memory pipeline with three stages.

\textbf{Stage 1: Context-guided interpretation.}
Rather than processing images in isolation, \textsc{VisualMem} interprets each image jointly with its surrounding conversational context. Language provides the disambiguating signal that raw visual content alone cannot supply. For example, ``this is my kitchen'' establishes user ownership, whereas ``I am visiting Marcus'' shifts attribution to a third party. Through this joint reasoning, the system resolves ambiguities in identity and scene ownership, producing a contextualized interpretation that identifies who is present, whose space is depicted, and which elements are plausible memory candidates.

\textbf{Stage 2: Deferred commitment.}
Grounding does not always succeed immediately. When the available evidence is insufficient for reliable interpretation, \textsc{VisualMem} places the image in a \emph{pending} state together with its original dialogue context, deferring any structured extraction. As the conversation evolves, a pending image is selectively re-evaluated when newly confirmed memory provides a relevant person or location reference. Rather than repeatedly scanning the entire pending set, the system revisits only images associated with that entity or scene, conditioning on the image, its original context, and the current memory state. Once the uncertainty blocking extraction is resolved, the image transitions to \emph{confirmed} and becomes eligible for extraction. This deferred commitment mechanism enables \textsc{VisualMem} to exploit long-range conversational dependencies and recover information that would otherwise be lost due to early uncertainty.

\textbf{Stage 3: Structured extraction.}
Confirmed images undergo structured extraction at three levels. At the \emph{relationship level}, the system infers social links and ownership associations. At the \emph{entity level}, it identifies recurring people, objects, and pets, storing visual references for future matching. At the \emph{fact level}, it extracts durable, visually grounded user facts (e.g., possessions or habits). These facts are verbalized and forwarded to the text-memory backend, which handles consolidation, deduplication, and temporal updates, integrating visual knowledge into the unified memory stream.

\subsection{Retrieval at Inference Time}
At query time, \textsc{VisualMem} routes the user's question to the most relevant memory source. Entity-centric queries, such as identifying a recurring person or recalling a user-owned object, are resolved against the structured visual memory store, which supports lookup by entity identity and fact type. Queries requiring broader factual, temporal, or preferential context are routed to the text-memory backend, or answered by combining evidence from both sources. This strategy allows \textsc{VisualMem} to complement rather than replace existing text-memory systems, with each memory source contributing the information it is best suited to provide.

\section{Experiments}
\subsection{Experimental Setup}
\textbf{Datasets.}
We evaluate our method on two benchmark groups. The first is our proposed \emph{personal visual memory benchmark}, where later questions depend on visually grounded information from earlier multimodal interactions. We report results on four settings: \emph{Target Person}, \emph{Target Asset}, \emph{Implicit Fact (Visual-Only)}, and \emph{Implicit Fact (Multimodal)}. The second group consists of established \emph{text-centric} long-term memory benchmarks, LOCOMO~\cite{locomo} and PersonaMem~\cite{personamem}, used to verify that improving visual memory does not degrade standard text-memory performance.

\textbf{Baselines.}
Following prior memory-system evaluations, we compare against three categories of baselines. \emph{Naive LLM} baselines include Full Context, which provides the entire interaction history at question time, and Oracle, which provides gold supporting evidence as an upper bound. \emph{RAG-based methods} include Self-RAG~\cite{selfrag} and HippoRAG2~\cite{hipporag2}. \emph{Memory-based methods} include LightMem~\cite{lightmem}, SimpleMem~\cite{simplemem}, Mem0~\cite{mem0}, and MemOS~\cite{memos}. Since \textsc{VisualMem} uses MemOS as its default text-memory backend, we compare directly with MemOS on text-centric benchmarks to measure whether the visual component preserves text-memory capability.

\textbf{Evaluation Metrics.}
We evaluate all settings under a multiple-choice protocol and report \emph{accuracy} as the main metric. We also report \emph{Tokens}, the amount of memory content provided to the answering model at inference time, to compare accuracy and retrieval efficiency.

\textbf{Implementation Details.}
\textsc{VisualMem} uses MemOS as the default text-memory backend and augments it with the proposed structured visual-memory module. On the visual benchmark, Gemini-Pro-3.1 is used for all methods during memory construction and question answering, allowing us to isolate the effect of the memory mechanism rather than the base model. On text-centric benchmarks, we follow the standard text-only protocol from prior work \cite{memos} and evaluate with GPT-4o-mini. Appendix~\ref{app:implementation} provides the complete implementation and retrieval configuration.

\subsection{Results on the Visual Memory Benchmark}
\begin{table}[t]
    \centering
    \renewcommand{\arraystretch}{1.25} 
    \caption{\textbf{Evaluation results on our visual memory benchmark.} We report performance for four task categories: \emph{Target Person}, \emph{Target Asset}, \emph{Implicit Fact (Visual-Only)}, and \emph{Implicit Fact (Multimodal)}. Each category is evaluated under a multiple-choice setting, and we report accuracy. Tokens denotes the amount of memory content provided to the answering model at inference time.}
    \vspace{1mm}
    \label{tab:evaluation}
    \resizebox{0.9\textwidth}{!}{%
    \begin{tabular}{@{} l c c c c c c @{}} 
    \toprule
    
    \textbf{Method} & \textbf{Tokens} & \textbf{Target Person} $\uparrow$ & \textbf{Target Asset} $\uparrow$ & \makecell{\textbf{Implicit Fact} \\ \textbf{(Visual-Only)} $\uparrow$} & \makecell{\textbf{Implicit Fact} \\ \textbf{(Multimodal)} $\uparrow$} & \textbf{Overall $\uparrow$} \\
    \midrule
    
    \multicolumn{7}{c}{\cellcolor{gray!15}\textbf{\textsc{Naive LLM}}} \\
    \midrule
    Full context    & 325K & 100.0 & 94.6 & 91.4 & 98.0 & 95.1 \\
    Oracle          & 1900 & 100.0 & 99.1 & 97.9 & 98.5 & 98.6 \\
    
    \midrule
    \multicolumn{7}{c}{\cellcolor{gray!15}\textbf{\textsc{RAG-Based Methods}}} \\
    \midrule
    Self-RAG \cite{selfrag}        & 2000 & 21.0 & 35.0 & 17.1 & 22.1 & 22.1 \\
    HippoRAG2 \cite{hipporag2}       & 1000 & 25.0 & 26.5 & 25.4 & 32.7 & 27.6 \\
    
    \midrule
    \multicolumn{7}{c}{\cellcolor{gray!15}\textbf{\textsc{Memory-Based Methods}}} \\
    \midrule
    LightMem \cite{lightmem}        & 500 & 30.0 & 40.2 & 45.4 & 58.8 & 46.1 \\
    SimpleMem \cite{simplemem}       & 500 & 3.0 & 40.2 & 42.9 & 43.2 & 36.8 \\
    Mem0  \cite{mem0}          & 500 & 38.0 & 33.9 & 40.8 & 57.9 & 45.0 \\
    MemOS  \cite{memos}        & 1187 & 45.0 & 59.9 & 52.1 & 64.8 & 56.0\\
    \rowcolor{blue!20} \textbf{\textsc{VisualMem} (Ours)} & 1980 & 95.0 & 91.1 & 77.9 & 83.4 & 84.1 \\
    \bottomrule
    \end{tabular}%
    }
\end{table}
Table~\ref{tab:evaluation} reports results on our visual memory benchmark. We first examine the two reference settings, Full Context and Oracle, to clarify the difficulty of the task. Full Context provides the entire interaction history to the model at question time, giving it direct access to all prior multimodal evidence but requiring a very large input budget. Oracle provides only the gold supporting evidence, and therefore serves as an upper bound on performance when the relevant memories are perfectly retrieved. The gap between Oracle and Full Context shows that simply placing the complete history in the context is not sufficient: the model must still identify the relevant visual evidence and reason over it among many distractors.

Practical memory systems face the harder problem of storing and retrieving compact memories without gold evidence. Existing RAG and memory baselines perform substantially worse than the reference settings, indicating that standard text-oriented memory mechanisms do not reliably preserve the visual information needed for later questions. \textsc{VisualMem} achieves the best performance among practical systems, with an overall accuracy of \textbf{84.1}, compared with 56.0 for the strongest memory-based baseline, MemOS. The largest gains appear on \emph{Target Person} and \emph{Target Asset}, where the system must track recurring people and user-associated objects across interactions. These cases are difficult for caption-based memory because local captions often fail to preserve identity, recurrence, ownership, or user relevance over time. \textsc{VisualMem} also improves performance on both \emph{Implicit Fact} settings, showing that its benefit is not limited to explicit entity recognition. These tasks require the system to infer implicit personal facts from visual evidence alone or from visual evidence combined with dialogue context. The gains suggest that \textsc{VisualMem} benefits from jointly reasoning over image content and dialogue context, storing structured visual memories, and keeping recurring entities and durable user facts accessible for later retrieval.

\subsection{Results on Text-Centric Benchmarks}
\begin{wraptable}{r}{0.48\textwidth}
\vspace{-1.4em}
\centering
\renewcommand{\arraystretch}{1.2}
\caption{\textbf{Evaluation results on long-term text-memory benchmarks.}}
\label{tab:text_benchmark}
\resizebox{0.48\textwidth}{!}{%
\begin{tabular}{l|cc}
\hline
Method & LOCOMO \cite{locomo} $\uparrow$ & PersonaMem \cite{personamem} $\uparrow$ \\
\hline
MemOS \cite{memos} & 56.8 & 45.5 \\
\rowcolor{blue!20} \textbf{\textsc{VisualMem} (Ours)} & 58.1 & 46.3 \\
\hline
\end{tabular}%
}
\vspace{-1em}
\end{wraptable}
Table~\ref{tab:text_benchmark} evaluates \textsc{VisualMem} on the text-centric LOCOMO and PersonaMem benchmarks, where the proposed visual-memory module is not required. Since \textsc{VisualMem} uses MemOS as its text-memory backend, we compare directly against MemOS to assess whether adding visual memory affects text-only performance. The results show comparable accuracy on both benchmarks, with small differences attributable to stochasticity in LLM-based memory construction and answering. This suggests that \textsc{VisualMem} preserves the underlying text-memory capability while extending long-term memory to visually grounded personal evidence.

\FloatBarrier
\subsection{Ablation Study}
\begin{table}[t]
\centering
\caption{\textbf{Ablation study of \textsc{VisualMem} on the visual memory benchmark.} Text, Visual, and Pending indicate whether the MemOS text-memory backend, structured visual memory, and deferred commitment for unresolved images are enabled. Window denotes the context size, with Full using the entire conversation session.}
\label{tab:ablation}
\vspace{1mm}
\resizebox{\textwidth}{!}{%
\begin{tabular}{lccccccccc}
\toprule
\textbf{Text} & \textbf{Visual} & \textbf{Pending} & \textbf{Window} & \textbf{Tokens} & \textbf{Target Person} $\uparrow$ & \textbf{Target Asset} $\uparrow$ & \makecell{\textbf{Implicit Fact} \\ \textbf{(Visual-Only)} $\uparrow$} & \makecell{\textbf{Implicit Fact} \\ \textbf{(Multimodal)} $\uparrow$} & \textbf{Overall} $\uparrow$ \\
\midrule
\checkmark & \checkmark &  & 2    & 1901 & 60.0 & 90.2 & 79.6 & 74.2 & 76.9 \\
\checkmark &  & \checkmark  & 2   & 1247 & 40.4 & 65.2 & 60.0 & 67.3 & 60.1 \\
 & \checkmark & \checkmark & 2   & 635 & 95.0 & 91.1 & 80.3 & 68.3 & 80.7 \\
\checkmark & \checkmark & \checkmark & 2    & 1882 & 95.0 & 91.1 & 76.1 & 76.9 & 81.5 \\
\checkmark & \checkmark & \checkmark & Full & 1980 & 95.0 & 91.1 & 77.9 & 83.4 & 84.1 \\
\bottomrule
\end{tabular}%
}
\vspace{1mm}
\end{table}

Table~\ref{tab:ablation} analyzes the main design choices in \textsc{VisualMem}. We examine the contributions of memory modality, deferred commitment, and conversational context, followed by controlled analyses of memory budget and hard-negative robustness.

\textbf{Memory modalities.}
Using only text memory performs substantially worse than using only visual memory, especially on \emph{Target Person} and \emph{Target Asset}. This confirms that the benchmark is primarily driven by visual evidence rather than by textual conversation history alone. Adding text memory on top of visual memory further improves performance, especially on \emph{Implicit Fact (Multimodal)}, where the answer depends on linking image evidence with dialogue context.

\textbf{Deferred commitment.}
Removing the pending mechanism causes a noticeable performance drop, particularly on \emph{Target Person}. This suggests that some visual memories are ambiguous when first observed, and committing to a final extraction too early can introduce noisy or incorrect entries. By deferring uncertain cases and revisiting them when additional visual context becomes available, \textsc{VisualMem} improves memory quality over time.

\textbf{Context window.}
We vary the number of neighboring dialogue turns provided with an image during memory construction. Using the full conversation session gives the best overall results, while restricting the window to size 2 weakens performance, especially on \emph{Implicit Fact (Multimodal)}. This indicates that broader context is important for linking visual evidence with textual cues when forming user facts.

\begin{figure}[t]
    \centering
    \begin{minipage}[t]{0.40\textwidth}
        \centering
        \captionof{table}{\textbf{Matched-budget comparison.} Overall accuracy with approximately matched inference-time memory budgets.}
        \label{tab:matched_budget}
        \vspace{1mm}
        \small
        \resizebox{\textwidth}{!}{%
        \begin{tabular}{lcc}
            \toprule
            \textbf{Token Budget} & \textbf{MemOS} & \textbf{\textsc{VisualMem}} \\
            \midrule
            $\sim$500  & 54.89 & 81.90 \\
            $\sim$1,000 & 55.89 & 83.76 \\
            $\sim$2,000 & 55.17 & 83.05 \\
            \bottomrule
        \end{tabular}%
        }
    \end{minipage}
    \hfill
    \begin{minipage}[t]{0.56\textwidth}
        \centering
        \captionof{table}{\textbf{Hard-negative verification.} Results on 438 known hard negatives. A match is a false positive, while rejection is the desired response.}
        \label{tab:hard_negative}
        \vspace{1mm}
        \small
        \resizebox{\textwidth}{!}{%
        \begin{tabular}{lccc}
            \toprule
            \textbf{Method} & \textbf{False Match $\downarrow$} & \textbf{Abstain $\downarrow$} & \textbf{Correct Rejection $\uparrow$} \\
            \midrule
            MemOS \cite{memos} & 0.68 & 56.85 & 42.47 \\
            \rowcolor{blue!20} \textbf{\textsc{VisualMem} (Ours)} & 1.14 & 3.20 & 95.66 \\
            \bottomrule
        \end{tabular}%
        }
    \end{minipage}
    \vspace{-4mm}
\end{figure}

\textbf{Matched memory budgets.}
To separate the effect of the memory architecture from the amount of retrieved content, we compare \textsc{VisualMem} and MemOS under approximately matched budgets of 500, 1,000, and 2,000 tokens. As shown in Table~\ref{tab:matched_budget}, \textsc{VisualMem} consistently outperforms MemOS at every budget by approximately 28 percentage points. This indicates that the improvement comes from preserving and retrieving structured visual information rather than from a larger inference-time memory budget.

\textbf{Hard-negative verification.}
We further test whether the systems incorrectly associate visually similar distractors with previously remembered entities. We convert all 438 hard-negative examples into direct identity-verification probes. Given a candidate image and its retrieved memory, the system predicts \emph{match} if the candidate is the same remembered entity, \emph{reject} if it is a different entity, or \emph{abstain} if the available evidence is insufficient to decide. Table~\ref{tab:hard_negative} shows that \textsc{VisualMem} correctly rejects 95.66\% of hard negatives while maintaining a 1.14\% false-match rate. MemOS has a similarly low false-match rate of 0.68\%, but abstains on 56.85\% of examples and explicitly rejects only 42.47\%. Thus, \textsc{VisualMem} more reliably distinguishes recurring entities from visually similar non-targets without relying on abstention.

\section{Conclusion}
We introduced \textsc{VisualMem}, a visual memory benchmark and architecture for long-term AI agents. Our benchmark evaluates whether models can retain visually grounded information from earlier multimodal interactions, including implicit facts and recurring visual entities. To address this setting, \textsc{VisualMem} augments a text-memory backend with structured visual memory that reasons over images and dialogue context and explicitly stores user-specific visual information. Experiments show that \textsc{VisualMem} substantially improves performance on our visual memory benchmark while remaining competitive on standard text-centric benchmarks. These results suggest that long-term multimodal memory requires more than generic caption storage: it needs dedicated mechanisms for preserving identity, ownership, and user-specific facts over time.

\section*{Limitations}
Our study has several limitations. First, the benchmark is synthetically constructed and contains only ten personas. While this design provides control over visual consistency and question grounding, it may not capture the full variability and ambiguity of naturally occurring longitudinal user histories. Second, parts of the pipeline rely on strong multimodal models for generation and interpretation, so benchmark quality and system performance may depend on the capabilities and biases of these underlying models. Third, we evaluate \textsc{VisualMem} on a limited number of personas and benchmarks, and do not study robustness under distribution shift, noisy images, or lower-quality visual inputs. Future work should extend the evaluation to privacy-preserving real-world interactions and more diverse multimodal settings.

\section*{Acknowledgements and Disclosures}
This work received no external funding. The authors declare no competing interests.

\bibliographystyle{unsrt}
\bibliography{neurips_2026}

\begin{thebibliography}{10}

\bibitem{locomo}
Adyasha Maharana, Dong-Ho Lee, Sergey Tulyakov, Mohit Bansal, Francesco
  Barbieri, and Yuwei Fang.
\newblock Evaluating very long-term conversational memory of llm agents.
\newblock In {\em Proceedings of the 62nd Annual Meeting of the Association for
  Computational Linguistics (Volume 1: Long Papers)}, pages 13851--13870, 2024.

\bibitem{longmemeval}
Di~Wu, Hongwei Wang, Wenhao Yu, Yuwei Zhang, Kai-Wei Chang, and Dong Yu.
\newblock Longmemeval: Benchmarking chat assistants on long-term interactive
  memory.
\newblock {\em arXiv preprint arXiv:2410.10813}, 2024.

\bibitem{personamem2}
Bowen Jiang, Yuan Yuan, Maohao Shen, Zhuoqun Hao, Zhangchen Xu, Zichen Chen,
  Ziyi Liu, Anvesh~Rao Vijjini, Jiashu He, Hanchao Yu, et~al.
\newblock Personamem-v2: Towards personalized intelligence via learning
  implicit user personas and agentic memory.
\newblock {\em arXiv preprint arXiv:2512.06688}, 2025.

\bibitem{mem0}
Prateek Chhikara, Dev Khant, Saket Aryan, Taranjeet Singh, and Deshraj Yadav.
\newblock Mem0: Building production-ready ai agents with scalable long-term
  memory.
\newblock {\em arXiv preprint arXiv:2504.19413}, 2025.

\bibitem{memos}
Zhiyu Li, Chenyang Xi, Chunyu Li, Ding Chen, Boyu Chen, Shichao Song, Simin
  Niu, Hanyu Wang, Jiawei Yang, Chen Tang, et~al.
\newblock Memos: A memory os for ai system.
\newblock {\em arXiv preprint arXiv:2507.03724}, 2025.

\bibitem{lee2024towards}
Eunhae Lee.
\newblock Towards ethical personal ai applications: Practical considerations
  for ai assistants with long-term memory.
\newblock {\em arXiv preprint arXiv:2409.11192}, 2024.

\bibitem{tomekcce2024private}
Batuhan T{\"o}mek{\c{c}}e, Mark Vero, Robin Staab, and Martin Vechev.
\newblock Private attribute inference from images with vision-language models.
\newblock {\em Advances in Neural Information Processing Systems},
  37:103619--103651, 2024.

\bibitem{he2024imagineyourself}
Zecheng He, Bo~Sun, Felix Juefei-Xu, Haoyu Ma, Ankit Ramchandani, Vincent
  Cheung, Siddharth Shah, Anmol Kalia, Harihar Subramanyam, Alireza Zareian,
  et~al.
\newblock Imagine yourself: Tuning-free personalized image generation.
\newblock {\em arXiv preprint arXiv:2409.13346}, 2024.

\bibitem{sundaram2024personalized}
Shobhita Sundaram, Julia Chae, Yonglong Tian, Sara Beery, and Phillip Isola.
\newblock Personalized representation from personalized generation.
\newblock {\em ArXiv}, abs/2412.16156, 2024.

\bibitem{peng2025dreambenchpp}
Yuang Peng, Yuxin Cui, Haomiao Tang, Zekun Qi, Runpei Dong, Jing Bai, Chunrui
  Han, Zheng Ge, Xiangyu Zhang, and Shu-Tao Xia.
\newblock Dreambench++: A human-aligned benchmark for personalized image
  generation.
\newblock {\em arXiv preprint arXiv:2406.16855}, 2024.

\bibitem{memorybank}
Wanjun Zhong, Lianghong Guo, Qiqi Gao, He~Ye, and Yanlin Wang.
\newblock Memorybank: Enhancing large language models with long-term memory.
\newblock In {\em Proceedings of the AAAI conference on artificial
  intelligence}, volume~38, pages 19724--19731, 2024.

\bibitem{perltqa}
Yiming Du, Hongru Wang, Zhengyi Zhao, Bin Liang, Baojun Wang, Wanjun Zhong,
  Zezhong Wang, and Kam-Fai Wong.
\newblock Perltqa: A personal long-term memory dataset for memory
  classification, retrieval, and fusion in question answering.
\newblock In {\em Proceedings of the 10th SIGHAN Workshop on Chinese Language
  Processing (SIGHAN-10)}, pages 152--164, 2024.

\bibitem{wang2024needle}
Weiyun Wang, Shuibo Zhang, Yiming Ren, Yuchen Duan, Tiantong Li, Shuo Liu,
  Mengkang Hu, Zhe Chen, Kaipeng Zhang, Lewei Lu, et~al.
\newblock Needle in a multimodal haystack.
\newblock {\em Advances in Neural Information Processing Systems},
  37:20540--20565, 2024.

\bibitem{wang2025multimodal}
Hengyi Wang, Haizhou Shi, Shiwei Tan, Weiyi Qin, Wenyuan Wang, Tunyu Zhang,
  Akshay Nambi, Tanuja Ganu, and Hao Wang.
\newblock Multimodal needle in a haystack: Benchmarking long-context capability
  of multimodal large language models.
\newblock In {\em Proceedings of the 2025 Conference of the Nations of the
  Americas Chapter of the Association for Computational Linguistics: Human
  Language Technologies (Volume 1: Long Papers)}, pages 3221--3241, 2025.

\bibitem{wu2024visual}
Tsung-Han Wu, Giscard Biamby, Jerome Quenum, Ritwik Gupta, Joseph~E Gonzalez,
  Trevor Darrell, and David~M Chan.
\newblock Visual haystacks: A vision-centric needle-in-a-haystack benchmark.
\newblock {\em arXiv preprint arXiv:2407.13766}, 2024.

\bibitem{shuster2021retrieval}
Kurt Shuster, Spencer Poff, Moya Chen, Douwe Kiela, and Jason Weston.
\newblock Retrieval augmentation reduces hallucination in conversation.
\newblock In {\em Findings of the Association for Computational Linguistics:
  EMNLP 2021}, pages 3784--3803, 2021.

\bibitem{ram2023incontext}
Ori Ram, Yoav Levine, Itay Dalmedigos, Dor Muhlgay, Amnon Shashua, Kevin
  Leyton-Brown, and Yoav Shoham.
\newblock In-context retrieval-augmented language models.
\newblock {\em Transactions of the Association for Computational Linguistics},
  11:1316--1331, 2023.

\bibitem{shi2024replug}
Weijia Shi, Sewon Min, Michihiro Yasunaga, Minjoon Seo, Richard James, Mike
  Lewis, Luke Zettlemoyer, and Wen-tau Yih.
\newblock Replug: Retrieval-augmented black-box language models.
\newblock In {\em Proceedings of the 2024 Conference of the North American
  Chapter of the Association for Computational Linguistics: Human Language
  Technologies (Volume 1: Long Papers)}, pages 8371--8384, 2024.

\bibitem{meminsight}
Rana Salama, Jason Cai, Michelle Yuan, Anna Currey, Monica Sunkara, Yi~Zhang,
  and Yassine Benajiba.
\newblock Meminsight: Autonomous memory augmentation for llm agents.
\newblock In {\em Proceedings of the 2025 Conference on Empirical Methods in
  Natural Language Processing}, pages 33124--33140, 2025.

\bibitem{memverse}
Junming Liu, Yifei Sun, Weihua Cheng, Haodong Lei, Yirong Chen, Licheng Wen,
  Xuemeng Yang, Daocheng Fu, Pinlong Cai, Nianchen Deng, et~al.
\newblock Memverse: Multimodal memory for lifelong learning agents.
\newblock {\em arXiv preprint arXiv:2512.03627}, 2025.

\bibitem{lightmem}
Jizhan Fang, Xinle Deng, Haoming Xu, Ziyan Jiang, Yuqi Tang, Ziwen Xu, Shumin
  Deng, Yunzhi Yao, Mengru Wang, Shuofei Qiao, et~al.
\newblock Lightmem: Lightweight and efficient memory-augmented generation.
\newblock {\em arXiv preprint arXiv:2510.18866}, 2025.

\bibitem{simplemem}
Jiaqi Liu, Yaofeng Su, Peng Xia, Siwei Han, Zeyu Zheng, Cihang Xie, Mingyu
  Ding, and Huaxiu Yao.
\newblock Simplemem: Efficient lifelong memory for llm agents.
\newblock {\em arXiv preprint arXiv:2601.02553}, 2026.

\bibitem{memgpt}
Charles Packer, Vivian Fang, Shishir\_G Patil, Kevin Lin, Sarah Wooders, and
  Joseph\_E Gonzalez.
\newblock Memgpt: towards llms as operating systems.
\newblock 2023.

\bibitem{amem}
Wujiang Xu, Zujie Liang, Kai Mei, Hang Gao, Juntao Tan, and Yongfeng Zhang.
\newblock A-mem: Agentic memory for llm agents.
\newblock {\em arXiv preprint arXiv:2502.12110}, 2025.

\bibitem{selfrag}
Akari Asai, Zeqiu Wu, Yizhong Wang, Avirup Sil, and Hannaneh Hajishirzi.
\newblock Self-rag: Learning to retrieve, generate, and critique through
  self-reflection.
\newblock In {\em The Twelfth International Conference on Learning
  Representations}, 2023.

\bibitem{hipporag2}
Bernal~Jim{\'e}nez Guti{\'e}rrez, Yiheng Shu, Weijian Qi, Sizhe Zhou, and
  Yu~Su.
\newblock From rag to memory: Non-parametric continual learning for large
  language models.
\newblock {\em arXiv preprint arXiv:2502.14802}, 2025.

\bibitem{wu2024personalized}
Junda Wu, Hanjia Lyu, Yu~Xia, Zhehao Zhang, Joe Barrow, Ishita Kumar, Mehrnoosh
  Mirtaheri, Hongjie Chen, Ryan~A Rossi, Franck Dernoncourt, et~al.
\newblock Personalized multimodal large language models: A survey.
\newblock {\em arXiv preprint arXiv:2412.02142}, 2024.

\bibitem{myvlm}
Yuval Alaluf, Elad Richardson, Sergey Tulyakov, Kfir Aberman, and Daniel
  Cohen-Or.
\newblock Myvlm: Personalizing vlms for user-specific queries.
\newblock In {\em European Conference on Computer Vision}, pages 73--91.
  Springer, 2024.

\bibitem{yollava}
Thao Nguyen, Haotian Liu, Yuheng Li, Mu~Cai, Utkarsh Ojha, and Yong~Jae Lee.
\newblock Yo'llava: Your personalized language and vision assistant.
\newblock {\em Advances in Neural Information Processing Systems},
  37:40913--40951, 2024.

\bibitem{yochameleon}
Thao Nguyen, Krishna~Kumar Singh, Jing Shi, Trung Bui, Yong~Jae Lee, and Yuheng
  Li.
\newblock Yo'chameleon: Personalized vision and language generation.
\newblock {\em 2025 IEEE/CVF Conference on Computer Vision and Pattern
  Recognition (CVPR)}, 2025.

\bibitem{rap}
Haoran Hao, Jiaming Han, Changsheng Li, Yu-Feng Li, and Xiangyu Yue.
\newblock Rap: Retrieval-augmented personalization for multimodal large
  language models.
\newblock In {\em Proceedings of the Computer Vision and Pattern Recognition
  Conference}, pages 14538--14548, 2025.

\bibitem{r2p}
Deepayan Das, Davide Talon, Yiming Wang, Massimiliano Mancini, and Elisa Ricci.
\newblock Training-free personalization via retrieval and reasoning on
  fingerprints.
\newblock In {\em Proceedings of the IEEE/CVF International Conference on
  Computer Vision}, pages 9683--9692, 2025.

\bibitem{personavlm}
Chang Nie, Chaoyou Fu, Yifan Zhang, Haihua Yang, and Caifeng Shan.
\newblock Personavlm: Long-term personalized multimodal llms.
\newblock {\em arXiv preprint arXiv:2604.13074}, 2026.

\bibitem{tame}
Rongpei Hong, Jian Lang, Ting Zhong, Yong Wang, and Fan Zhou.
\newblock Tameing long contexts in personalization: Towards training-free and
  state-aware mllm personalized assistant.
\newblock In {\em Proceedings of the 32nd ACM SIGKDD Conference on Knowledge
  Discovery and Data Mining V. 1}, pages 452--463, 2026.

\bibitem{personahub}
T~Ge, X~Chan, X~Wang, D~Yu, H~Mi, and D~Yu.
\newblock Scaling synthetic data creation with 1,000,000,000 personas. arxiv
  2024.
\newblock {\em arXiv preprint arXiv:2406.20094}.

\bibitem{personamem}
Bowen Jiang, Zhuoqun Hao, Young-Min Cho, Bryan Li, Yuan Yuan, Sihao Chen, Lyle
  Ungar, Camillo~J Taylor, and Dan Roth.
\newblock Know me, respond to me: Benchmarking llms for dynamic user profiling
  and personalized responses at scale.
\newblock {\em arXiv preprint arXiv:2504.14225}, 2025.

\end{thebibliography}
\clearpage
\appendix
\tcbset{
  promptbox/.style={
    enhanced,
    breakable,
    width=\textwidth,
    enlarge left by=0mm,
    enlarge right by=0mm,
    colback=blue!5,
    colframe=blue!70!black,
    coltitle=white,
    colbacktitle=blue!70!black,
    fonttitle=\small\bfseries,
    arc=1mm,
    boxrule=0.7pt,
    left=1mm,
    right=1mm,
    top=0.5mm,
    bottom=0.5mm,
    boxsep=0.5mm,
    toptitle=0.5mm,
    bottomtitle=0.5mm
  },
  promptoutputbox/.style={
    enhanced,
    width=\linewidth,
    colback=white,
    colframe=green!35!black,
    boxrule=0.5pt,
    arc=1mm,
    left=0.5mm,
    right=0.5mm,
    top=0.5mm,
    bottom=0.5mm,
    boxsep=0.5mm
  }
}

\lstdefinestyle{promptlisting}{
  basicstyle=\ttfamily\footnotesize,
  breaklines=true,
  breakatwhitespace=true,
  breakindent=0pt,
  breakautoindent=false,
  escapeinside={(*@}{@*)},
  columns=fullflexible,
  keepspaces=true,
  showstringspaces=false
}

\section{\textsc{VisualMem} Implementation Details}
\label{app:implementation}

\textsc{VisualMem} is implemented as a wrapper around MemOS, which handles the text-memory stream. We use \texttt{gemini-3.1-pro-preview} for context routing, structured extraction, scene comparison, question parsing, and answer generation. No additional model is trained. Events are processed chronologically and separately for each persona. Unless stated otherwise, the \texttt{window-full} setting provides an image with the full conversation from its event and the event date. Each persona is assigned a JSON memory file and a Qdrant collection. The JSON file contains the user profile, relationships, objects, pets, private locations, confirmed images, and pending images.

\textbf{Memory construction.}
For an image turn, the routing model receives the image, its conversation context, and a list of people with available face references. The output specifies possible identities, scene ownership, and whether identity or location ambiguity prevents extraction. If the image can be committed, the extraction model records its scene summary, photo time, people, objects, pets, user facts, and retrieval tags. The model also assigns each visible face a visibility score between 0 and 10. For each person, we keep the reference image with the highest score observed so far. We embed the scene summary and tags using \texttt{gemini-embedding-2-preview} with 1,536 output dimensions, then store the vector in Qdrant with cosine distance. Durable user facts are converted to text and added to the corresponding conversation turn before it is sent to MemOS. The remaining entity and image metadata stays in the visual store.

\textbf{Pending images.}
Pending images are re-evaluated selectively when relevant new evidence becomes available, rather than at fixed intervals or after every image turn. An image remains pending when unresolved identity or scene information prevents reliable structured extraction, such as when a suitable face or location reference is unavailable. Identity-related images are reconsidered when a reliable face reference for a candidate person is obtained. Scene-related images are reconsidered when a relevant location reference becomes available. If the required reference is already stored, the comparison is performed immediately. The system does not repeatedly scan the full pending set. It revisits only images associated with the newly confirmed person or location. Once all uncertainty blocking extraction has been resolved, the image is re-analyzed using its original conversation context and the newly available references, then stored as confirmed structured memory. Across 1,718 image turns, 98 required identity or scene resolution, corresponding to approximately one pending case every 17.5 turns.

\textbf{Retrieval.}
The question parser returns semantic search terms, object names, pet terms, and an optional photo-time constraint. Objects and pets are matched against their normalized names and descriptions. For image retrieval, the semantic terms are embedded with the retrieval-query setting of \texttt{gemini-embedding-2-preview}. We request 10 results from Qdrant, combine them with exact metadata and time matches, and place the first three image summaries in the answering context. Retrieval is rank based and does not use a similarity threshold. The same question is sent to MemOS, which returns at most 20 text memories in fast-search mode and six preference records. The final prompt combines both memory sources with the question and its answer choices. When the question or a choice contains an image, that image is supplied directly to the answering model.

\section{Benchmark Validation and Quality Control}
\label{app:benchmark_validation}

\textbf{Implicit facts.}
We first screen implicit-fact examples with three independent VLM judges: Gemini Pro 3.1, GPT-5.2, and Claude Opus 4.6. Majority voting is used to remove examples in which the intended fact is ambiguous, overly specific, temporary rather than durable, unsupported by the evidence, or grounded in an image with visible generation artifacts. The remaining examples are reviewed independently by three internal researchers under the same criteria. Retention requires unanimous agreement that the fact is clearly and uniquely supported. For visual-only examples, the image must provide sufficient evidence on its own. For multimodal examples, the fact must follow from the image and dialogue together. Approximately 28\% of the initially generated implicit-fact conversations pass both stages.

\textbf{Explicit entities and distractors.}
Three internal researchers review every conversation-image pair used for explicit-entity and distractor examples. For an explicit entity, the target person or asset must be visible, recognizable, consistent with its reference, and correctly associated with the conversation. For a distractor, the image must not imply user ownership, a known identity, or another benchmark-relevant fact. A pair is accepted only with unanimous agreement. Failed pairs are regenerated and reviewed again.

\textbf{Question validation.}
Questions are generated only from conversation-image pairs that pass the preceding checks. Three internal researchers then inspect the image, conversation, question, answer choices, and ground-truth answer as a single package. A question is retained only when all reviewers agree that the queried identity, attribute, or fact is present in the evidence; the ground-truth answer is uniquely supported; no alternative option is equally plausible; the question does not expose the answer through textual leakage; and the distractors are plausible but unsupported. Questions that fail any criterion are discarded. The final benchmark contains 696 questions that pass this review.

\section{Broader Impacts}
This work may have positive impact by improving long-term multimodal memory in personalized AI assistants, which could make such systems more helpful in settings such as accessibility support, daily assistance, and personalized human--AI interaction. At the same time, stronger visual memory also introduces risks: systems that better retain user-specific visual information could be misused for profiling, surveillance, or intrusive inference about private attributes from personal images. In addition, incorrect visual memory could lead to false inferences about identity, ownership, or personal facts. To reduce these risks, our benchmark is constructed synthetically rather than from real user histories, and future deployment of such systems should incorporate privacy safeguards, user control over stored memories, and careful restrictions on sensitive inference.

\section{Additional Experimental Results}

\subsection{Open-Ended Evaluation}
The main experiments use a multiple-choice protocol, which may introduce answer-choice artifacts. To evaluate whether the performance gains persist without access to candidate answers, we conduct an open-ended evaluation on the full benchmark. The answering model receives only the retrieved memory and the question, without answer choices. A separate Gemini evaluator maps each free-form response to its corresponding answer option without access to the retrieved memory or ground-truth label; ambiguous or unmatched responses are counted as incorrect.

As shown in Table~\ref{tab:open_ended}, \textsc{VisualMem} achieves 58.05\% open-ended accuracy, compared with 13.94\% for MemOS. \textsc{VisualMem} therefore remains substantially better than MemOS without answer choices, suggesting that its advantage does not depend on multiple-choice cues.
\begin{center}
\centering
\captionof{table}{\textbf{Open-ended evaluation.} The answering model receives the retrieved memory and question without answer choices.}
\label{tab:open_ended}
\vspace{1mm}
\begin{tabular}{lc}
\toprule
\textbf{Method} & \textbf{Open-Ended Accuracy $\uparrow$} \\
\midrule
MemOS & 13.94 \\
\rowcolor{blue!20} \textbf{\textsc{VisualMem} (Ours)} & 58.05 \\
\bottomrule
\end{tabular}
\vspace{-3mm}
\end{center}

\subsection{Multiple-Run Analysis}
We repeat the complete evaluation pipeline five times for Mem0, MemOS, and \textsc{VisualMem}. Each run rebuilds memory from the full interaction history before rerunning retrieval and answer generation for all questions. Table~\ref{tab:multiple_runs} reports mean accuracy and standard deviation. \textsc{VisualMem} reaches $84.11 \pm 0.87$ overall, compared with $56.08 \pm 0.82$ for MemOS and $44.97 \pm 0.81$ for Mem0, well beyond the observed run-to-run variation.
\begin{center}
\centering
\captionof{table}{\textbf{Multiple-run evaluation.} Mean accuracy $\pm$ standard deviation over five complete runs.}
\label{tab:multiple_runs}
\vspace{1mm}
\resizebox{\textwidth}{!}{%
\begin{tabular}{lccccc}
\toprule
\textbf{Method} & \textbf{Target Person $\uparrow$} & \textbf{Target Asset $\uparrow$} & \makecell{\textbf{Implicit Fact} \\ \textbf{(Visual-Only)} $\uparrow$} & \makecell{\textbf{Implicit Fact} \\ \textbf{(Multimodal)} $\uparrow$} & \textbf{Overall $\uparrow$} \\
\midrule
Mem0  & $38.40 \pm 2.07$ & $34.52 \pm 1.83$ & $41.52 \pm 2.11$ & $58.62 \pm 1.69$ & $44.97 \pm 0.81$ \\
MemOS & $45.40 \pm 2.07$ & $59.82 \pm 1.91$ & $52.08 \pm 1.73$ & $64.86 \pm 2.07$ & $56.08 \pm 0.82$ \\
\rowcolor{blue!20} \textbf{\textsc{VisualMem} (Ours)} & $95.20 \pm 1.92$ & $91.09 \pm 1.87$ & $77.83 \pm 2.21$ & $83.43 \pm 1.61$ & $84.11 \pm 0.87$ \\
\bottomrule
\end{tabular}%
}
\vspace{-4mm}
\end{center}

\section{Prompts Used by \textsc{VisualMem}}
The following prompts are used for visual-memory construction and retrieval. Red text marks inputs supplied at runtime.
\vspace{1mm}
\begin{center}
\begin{tcolorbox}[
    promptbox,
    title={\textbf{Analyzing context for visual-memory routing.}}
]
\begin{lstlisting}[
    style=promptlisting
]
Analyze the conversation context and image for visual-memory routing.
(*@{\rmfamily\bfseries Decide:}@*)
1. Who might need identity matching
2. Which face reference images should be loaded
3. Whose space/environment is shown
4. Whether extraction should wait in pending

You also receive the people who currently have face reference images in memory; use that real availability when deciding pending.

(*@{\rmfamily\bfseries Output format:}@*) Return ONLY this JSON object, with no markdown fences:
{
  "people_possibly_in_image": ["User", "Sarah"],
  "mentioned_people": [
    {"name": "Sarah", "relationship": "friend", "evidence": "user said 'Sarah and I'"}
  ],
  "load_all_known_faces": false,
  "user_explicitly_confirmed": false,
  "scene_ownership": "user_space|other_person_space|shared_space|public_space|unknown",
  "scene_owner_name": null,
  "private_location": {
    "name": "kitchen",
    "description": "short visual description of the user's private room"
  },
  "possible_location": ["kitchen"],
  "pending": false,
  "pending_types": [],
  "pending_reason": "",
  "reasoning": "..."
}

(*@{\rmfamily\bfseries Identity rules:}@*)
- "people_possibly_in_image": people who could physically be visible and may need identity resolution. Include "User" when the user could be in the image.
- "mentioned_people": every named person in context, even if not visible.
- "user_explicitly_confirmed": true only for explicit self-photo language such as "picture of me", "this is me", "selfie", or "photo of me doing X".
- "load_all_known_faces": true when unnamed companion/group language or the image itself implies unaccounted-for visible people. This includes "we", "our", "us", unnamed companions, or multiple main/foreground people when context names only some of them. Include known/strongly implied people such as "User", but do not invent names for unnamed companions.
- Add "identity" to "pending_types" only when multiple plausible visible identities remain and available face refs are insufficient. If exactly one named non-user person is the only plausible identity target, pending can be false via the caller's deterministic shortcut. If refs are sufficient, keep pending false and explain why.

(*@{\rmfamily\bfseries Scene ownership rules:}@*)
- "user_space": the user's private/user-owned space, such as "my kitchen", "my living room", "our bedroom", "my workspace", "my desk setup", or a room in the user's home.
- "other_person_space": another named person's private space, such as "Marcus's apartment", "his kitchen", "Maya's desk in our apartment", or "her workspace". Specific ownership beats shared-home wording. Set "scene_owner_name" when known and do not save "private_location".
- "shared_space": a genuinely shared private space not assigned to a specific person, such as "our kitchen" or "our living room".
- "public_space": restaurants, parks, stores, offices, gyms, streets, events, libraries, yards/gardens/outdoor areas, and other non-private places.
- "unknown": a possibly private room/space whose owner is not established. Add "scene" to "pending_types" in this case.

(*@{\rmfamily\bfseries Location fields:}@*)
- For "user_space", fill "private_location" with a strict indoor room/space name and a short visual description. Good names include "kitchen", "living room", "bedroom", "workspace", "dining room", "bathroom", "garage", "home office", "entryway", "hallway", and "laundry room". If needed, use another concise indoor room/space noun phrase.
- Never include ownership words such as "my", "our", "user's", or "the user's" in location names.
- "possible_location" is a list of strict indoor private room/space names that could match this scene, especially for "unknown" scenes that may later be compared to known user-owned locations.
- Do not use outdoor/public places or generic object/furniture/surface names as "private_location.name" or "possible_location": e.g. yard, garden, library, park, street, desk, table, counter, shelf, wall, couch, or bed.

(*@{\rmfamily\bfseries Close-up/surface exceptions:}@*)
- If the image is only a close-up of a desk, table, counter, shelf, wall, object, pet, document, screen, or small surface, and the broader room cannot be identified, do not scene-pend. Use "scene_ownership" = "unknown" if needed, but leave "scene" out of "pending_types" and set "possible_location" = [].
- If the table/desk/counter/surface is the main visible subject, treat it as an object/surface scene. Do not infer "workspace", "home office", "dining room", or similar unless the broader room is clearly visible or context explicitly says so.
- Do not infer "workspace" or "home office" merely from a desk, notes, laptop, monitor, paperwork, or someone working.

(*@{\rmfamily\bfseries Pending rules:}@*)
- "pending" must be true exactly when "pending_types" is non-empty.
- If "user_space" is clear but identity is unresolved, still provide "private_location"; the caller can save the space while leaving identity pending.
- If no identity matching is needed and scene ownership is not unknown, set pending false.
\end{lstlisting}
\end{tcolorbox}
\vspace{-2mm}
\captionof{figure}{Prompt template for analyzing conversation context and routing visual memories.}
\label{fig:prompt_visualmem_context_analysis}
\end{center}

\begin{center}
\begin{tcolorbox}[
    promptbox,
    title={\textbf{Matching unresolved scenes.}}
]
\begin{lstlisting}[
    style=promptlisting
]
Compare confirmed reference images with a pending image and decide whether they show the same user-owned private room or space.

(*@{\rmfamily\bfseries Inputs:}@*)
- 1-3 confirmed reference images of one user-owned private location
- 1 pending image with unresolved scene ownership/location
- Candidate location name and any known description

(*@{\rmfamily\bfseries Output format:}@*) Return ONLY this JSON object, with no markdown fences:
{
  "same_location": true,
  "confidence": 0.91,
  "location_name": "kitchen",
  "reasoning": "short explanation of matching visual cues"
}

(*@{\rmfamily\bfseries Rules:}@*)
- Set "same_location" = true only when stable visual cues strongly match at least one reference image: cabinets, counters, furniture layout, wall color, desk setup, windows, appliances, decor, or other distinctive fixed details.
- Do not match only because both images share a generic room type; two kitchens are different locations unless their details match.
- "location_name" must be a strict room/space noun phrase such as "kitchen", "living room", "bedroom", "workspace", or "dining room". Do not include ownership words such as "my", "our", "user's", or "the user's".
- If unsure, set "same_location" = false.
\end{lstlisting}
\end{tcolorbox}
\vspace{-2mm}
\captionof{figure}{Prompt template for matching a pending image against known private locations.}
\label{fig:prompt_visualmem_scene_match}
\end{center}

\begin{center}
\begin{tcolorbox}[
    promptbox,
    title={\textbf{Extracting structured visual memories.}}
]
\footnotesize
\textbf{System prompt}
\begin{lstlisting}[
    style=promptlisting
]
You are a visual-memory analyst. Given an image and conversation context, extract structured information. (*@{\rmfamily\bfseries Return ONLY a JSON object, with no markdown fences.}@*)

{
  "scene_summary": "Detailed description of the full scene using real names when known, with extra detail about main people.",
  "scene_type": "shopping|office|outdoor|home|event|social|travel|food|sport|other",
  "photo_time": "ISO date/time or date from context, e.g. 2024-01-14, otherwise null",
  "people_in_image": [
    {
      "label": "canonical name from mentioned_people if known, otherwise descriptive label",
      "is_user": false,
      "reasoning": "Why you think this is/isn't the user",
      "position": "left side, foreground",
      "face_visibility_score": 0-10
    }
  ],
  "can_confirm_user": true/false,
  "user_confirmation_reasoning": "Explain why user identity can or cannot be confirmed",
  "mentioned_people": [
    {"name": "Sarah", "relationship": "friend", "evidence": "user said 'Sarah and I'"}
  ],
  "objects_in_image": [
    {"name": "laptop", "owner": "user|person_name|unknown", "description": "silver MacBook"}
  ],
  "pets_in_image": [
    {"name": "Buddy", "type": "dog", "breed": "golden retriever", "owner": "user|person_name|unknown", "description": "wearing red collar"}
  ],
  "user_facts": [
    {"statement": "I am training for a half-marathon.", "evidence": ["User said '...half-marathon training'"], "confidence": 0.95}
  ],
  "tags": ["running", "shopping", "shoes", "Sarah"]
}

(*@{\rmfamily\bfseries User confirmation:}@*)
- Set "can_confirm_user" = true only for explicit self-photo context such as "picture of me", "this is me", "selfie", "photo of me doing X", "snapped this picture of me", or "picture of us".
- Set "can_confirm_user" = false when there is no explicit self-reference, or when wording like "Sarah and I are at X" plus one visible person remains ambiguous.
- When "can_confirm_user" = false, set every "people_in_image[].is_user" = false. The image may go to pending memory until identity is confirmed.

(*@{\rmfamily\bfseries People and names:}@*)
- Include only main/important visible people central to the image or context. Ignore background bystanders, crowd faces, reflections, posters/screens, and tiny or blurred distant people unless context asks about them.
- Do not match faces against known references; that is handled separately.
- Label visible people with the exact canonical "mentioned_people.name" when applicable. Otherwise use a descriptive label.
- Relationship/role words are not real names. For "my dad", "my mom", "my brother", "my neighbor", etc., use "name" = "unknown" and the relationship value. For "my dad Thomas" or "my mom Grace", use the real name and the relationship.
- Keep canonical names consistent across "people_in_image.label", owners, tags, and "scene_summary".
- "face_visibility_score" is 0-10 for usefulness as a face reference. Score high only for close, clear, mostly frontal, well-lit, unobstructed faces. Score low for distant, tiny, blurred, side/back views, masks, sunglasses, occlusions, or invisible faces. Be conservative when multiple people appear.

(*@{\rmfamily\bfseries Scene, time, objects, pets, and tags:}@*)
- "scene_summary": write a concrete full-scene description with setting, activity, important objects, and each main person's clothing, held items, posture, and action. Use real names when identity is known; use "User" consistently if the user is identified.
- "photo_time": extract the photo date/time from conversation context, including DATE lines. Use ISO-like strings such as "2024-01-14" or "2024-01-14 evening"; otherwise null. Do not put date/time in tags.
- "objects_in_image": include name, owner, and visual description. Use "owner" = "user" only for the user's possessions. Describe intrinsic visual details such as brand/model, color, size, material, condition, markings, stickers, case, or attached accessories. Do not describe location unless it is physically part of the object.
- "pets_in_image": include pet name/type/breed if known, owner, and visual details such as color, size, coat pattern, collar/harness/clothing, pose, expression, and distinctive markings. "His cat" means owner=person_name, not user. Do not describe surrounding furniture or room unless attached to or worn by the pet.
- "tags": provide 5-10 search keywords covering visible things, people names, activities, scene type, and location clues. Exclude photo date/time.

(*@{\rmfamily\bfseries User facts:}@*)
- Store only durable personal information about the user, written in first person starting with "I ...". Do not write "User ...".
- Facts must be grounded in the image alone or in the image plus user-provided conversation context. Use context only when it explains visible evidence, such as ownership, identity, activity, relationship, or durable preference.
- Never use assistant replies as evidence. Never store unrelated context-only claims that the image does not help ground.
- Do not store transient scene details such as "I am holding a coffee cup right now"; those belong in "scene_summary" or "objects_in_image".
- A valid fact should likely still be true tomorrow, e.g. "I own a silver MacBook Pro", "I am training for a half-marathon", or "I have a baby".
- Each fact needs a durable, self-contained first-person "statement", visual and/or user-context "evidence" with relevant user quote/paraphrase when used, and "confidence" from 0.0 to 1.0.

(*@{\rmfamily\bfseries User fact examples:}@*)
- Cat on the user's living room sofa -> "I own a cat."
- User's desk with MacBook, external monitor, and mechanical keyboard -> "I own a silver MacBook Pro." and "I use a dual-monitor setup for work."
- User at Mount Fuji with trail-running vest and race bib -> "I climb mountains / hike." and "I participate in trail-running races."
- User's kitchen with vegetables, tofu, and a vegan cookbook -> "I follow a vegan / plant-based diet."
- User's baby sleeping in a nursery -> "I have a baby / child."
- User in a white coat with stethoscope in a hospital ward -> "I work in healthcare (likely a doctor)."
- User's bookshelf filled with science-fiction novels -> "I enjoy reading science-fiction novels."
- User holding one coffee cup at a cafe -> no fact unless context provides a durable habit or preference.

Never invent facts the image cannot support. One coffee cup does not imply "I drink coffee every day" without repeated evidence or explicit user context.
\end{lstlisting}
\medskip
\textbf{User prompt}
\begin{lstlisting}[
    style=promptlisting
]
Analyze this image and extract all information per your instructions. (*@{\rmfamily\bfseries Return ONLY the JSON object.}@*)

(*@{\rmfamily\bfseries Conversation context:}@*) (*@\textcolor{red}{\{context\}}@*)

(*@\textcolor{red}{\{memory\_block\}}@*)
\end{lstlisting}
\end{tcolorbox}
\vspace{-2mm}
\captionof{figure}{System and user prompt templates for extracting structured visual memories.}
\label{fig:prompt_visualmem_extraction}
\end{center}

\begin{center}
\begin{tcolorbox}[
    promptbox,
    title={\textbf{Analyzing questions for memory retrieval.}}
]
\begin{lstlisting}[
    style=promptlisting
]
Analyze the user's question and choose what to search in visual memory.

(*@{\rmfamily\bfseries Memory includes:}@*)
- User profile: owned objects and pets
- Images: summaries, people/pets/objects, tags, and "photo_time"

(*@{\rmfamily\bfseries Output format:}@*) Return ONLY this JSON object, with no markdown fences:
{
  "search_keywords": ["keyword1", "keyword2"],
  "object_names": ["laptop", "notebook computer", "computer"],
  "pet_keywords": ["dog", "cat"],
  "photo_time": "2024-01-14 | November 2017 | null",
  "reasoning": "..."
}

(*@{\rmfamily\bfseries Rules:}@*)
- "search_keywords": 2-5 semantic terms for Qdrant, focused on scene, activity, location, people, or event clues.
- If the question is clearly about a specific object or pet, set "search_keywords" = [] and put those terms only in "object_names" or "pet_keywords"; do not duplicate object/pet terms in "search_keywords".
- "object_names": include the exact requested object plus generous visual and semantic variants: synonyms, everyday names, broader/narrower categories, compounds, brands/types, and nearby confusable names. Example: "book" can include book, notebook, textbook, journal, diary, planner, binder, paperback, hardcover, magazine, and manual; "laptop" can include laptop, notebook computer, computer, MacBook, Chromebook, and portable computer.
- "pet_keywords": pet-related terms, including species, breed/type, pet name, or generic terms like dog/cat when useful.
- "photo_time": the requested photo/event date or time constraint, otherwise null. For old events, use the event/photo date, not the current conversation date.
\end{lstlisting}
\end{tcolorbox}
\vspace{-2mm}
\captionof{figure}{Prompt template for translating a question into visual-memory retrieval fields.}
\label{fig:prompt_visualmem_question_analysis}
\end{center}

\clearpage
\section{Prompts Used in Our Dataset Generation}
The prompts below cover dataset construction and benchmark question generation. Red text marks inputs supplied at runtime.

\subsection{Persona and Timeline Construction}
\begin{center}
\begin{tcolorbox}[
    promptbox,
    title={\textbf{Generating social connections.}}
]
\footnotesize

\textbf{Role:} You are a Narrative Designer and Data Architect.\\
\textbf{Input Context:} \textcolor{red}{\{expanded\_persona\}}\\
\textbf{Task:} Create a list of 5 to 7 distinct characters who have a close daily life or professional connection to the main persona.

\vspace{0.2em}
\textbf{Requirements:}
\begin{enumerate}[leftmargin=1.2em, itemsep=0pt, topsep=0pt, parsep=0pt]
    \item \textbf{Rich Diversity \& Balance:} Do NOT fill the list with only coworkers. You must generate a heterogeneous social circle including:
    \begin{itemize}[leftmargin=1em, itemsep=0pt, topsep=0pt, parsep=0pt]
        \item \textbf{Family \& Romantic:} Parents, siblings, cousins, spouse/partner, or children.
        \item \textbf{Social \& Community:} Childhood friends, college roommates, neighbors, gym/hobby acquaintances.
        \item \textbf{Professional \& Academic:} Mentors, direct reports, investors, long-term clients, academic peers, or rivals.
    \end{itemize}
    \item \textbf{Realism:} Names and backgrounds must be culturally consistent with the main persona.
    \item \textbf{Visuals:} The \texttt{detailed\_identity} field must be a photorealistic image prompt describing physical appearance, outfit, and environment. Do not describe personality here. The person is looking towards the camera.
\end{enumerate}

\textbf{Output Format:} Return ONLY valid JSON.

\vspace{0.2em}
\begin{tcolorbox}[promptoutputbox]
\scriptsize
\begin{alltt}
\{
  "social_relationship": \{
    "\textcolor{red}{\{idx_persona\}}_0": \{
      "name": "[Name]",
      "description": "[Age, job, and why they interact with the main persona]",
      "detailed_identity": "Subject: [Ethnicity, gender, age, physical features]. 
                            Outfit: [Realistic daily or work attire]. 
                            Environment: [A specific real-world location].",
      "relationship": "[Short label, e.g., Sister, Best Friend]"
    \},
    "\textcolor{red}{\{idx_persona\}}_1": \{ ... \}
  \}
\}
\end{alltt}
\end{tcolorbox}

\end{tcolorbox}
\vspace{-2mm}
\captionof{figure}{Prompt template for generating recurring social connections.}
\label{fig:prompt_social_connections}
\end{center}

\begin{center}
\begin{tcolorbox}[
    promptbox,
    title={\textbf{Generating persona assets.}}
]
\footnotesize

\textbf{Role:} You are the Persona Asset Generator for a visual reasoning dataset.

\textbf{Task:} Based on the provided User Persona, generate a list of specific, persistent ``Objects'' (physical items) and ``Agents'' (pets or animals) that belong to or are closely associated with this user.

\textbf{Inputs:}
\begin{itemize}[leftmargin=1em, itemsep=0pt, topsep=0pt, parsep=0pt]
    \item \textbf{User Persona:} \textcolor{red}{\{persona\_json\}}
\end{itemize}

\textbf{Step-by-Step Generation Rules:}
\begin{enumerate}[leftmargin=1.2em, itemsep=0pt, topsep=0pt, parsep=0pt]
    \item \textbf{Analyze the Persona:} Review the user's occupation, hobbies, income, and living situation to determine what items or pets they would realistically own.
    \item \textbf{Brainstorm Assets:} Create a diverse list of 5 to 8 distinct assets.
    \begin{itemize}[leftmargin=1em, itemsep=0pt, topsep=0pt, parsep=0pt]
        \item \textbf{Objects:} Include items like specific vehicles, customized electronics, favorite accessories, or hobby equipment.
        \item \textbf{Agents:} Include pets, e.g., a specific breed of dog or a cat with distinct markings.
    \end{itemize}
    \item \textbf{Visual Specificity:} For every asset, provide a dense, hyper-realistic physical description. Include colors, textures, materials, size, and distinctive marks, e.g., scratches, stickers, or unique patterns, so it can be consistently rendered in text-to-image prompts.
\end{enumerate}

\textbf{Output Format:} Return ONLY valid JSON.

\vspace{0.2em}
\begin{tcolorbox}[promptoutputbox]
\scriptsize
\begin{alltt}
\{
  "assets": [
    \{
      "asset_id": "asset_01",
      "name": "[Short name, e.g., 'Whiskers the Calico Cat' or 'Sticker-Covered Laptop']",
      "visual_description": "[Dense visual description suitable for an image generator prompt]",
      "persona_connection": "[1-sentence reason why this persona owns this item/agent]"
    \}
    ... more asset
  ]
\}
\end{alltt}
\end{tcolorbox}

\end{tcolorbox}
\vspace{-2mm}
\captionof{figure}{Prompt template for generating persistent persona-associated assets.}
\label{fig:prompt_asset_generation}
\end{center}

\clearpage
\begin{center}
\begin{tcolorbox}[
    promptbox,
    title={\textbf{Generating event timelines.}}
]
\footnotesize

\textbf{Role:} You are the Narrative Logic Engine. Your goal is to simulate a realistic, causally linked timeline of events.

\textbf{Task:} Generate a timeline of exactly 5 to 8 distinct events based on the user-provided topic.

\vspace{0.2em}
\textbf{Inputs:}
\begin{itemize}[leftmargin=1em, itemsep=0pt, topsep=0pt, parsep=0pt]
    \item \textbf{User Persona:} \textcolor{red}{\{persona\_json\}}
    \item \textbf{Topic:} \textcolor{red}{\{topic\}}
    \item \textbf{Simulation Start Date:} \textcolor{red}{\{start\_date\}}
\end{itemize}

\textbf{Critical Constraints (Logic \& Causality):}
\begin{enumerate}[leftmargin=1.2em, itemsep=0pt, topsep=0pt, parsep=0pt]
    \item \textbf{Chronological Consistency:} Dates must progress linearly.
    \item \textbf{Unidirectional Causality:}
    \begin{itemize}[leftmargin=1em, itemsep=0pt, topsep=0pt, parsep=0pt]
        \item \textbf{Allowed:} An earlier event can cause a later event.
        \item \textbf{Allowed:} Events can be completely unrelated.
        \item \textbf{Forbidden:} A later event cannot cause an earlier event.
    \end{itemize}
    \item \textbf{Flexibility:} Not every event must be connected. If an event is standalone, use \texttt{"None"} for the causal link.
\end{enumerate}

\textbf{Output Format:} Return ONLY valid JSON.

\vspace{0.2em}
\begin{tcolorbox}[promptoutputbox]
\scriptsize
\begin{alltt}
\{
  "events": [
    \{
      "event_id": 1,
      "date": "YYYY-MM-DD",
      "summary": "[Concise description of the action]",
      "topic": "[Topic of the event]",
      "causal_link": "None"
    \},
    \{
      "event_id": 2,
      "date": "YYYY-MM-DD",
      "summary": "[Consequence or follow-up event]",
      "topic": "[Topic of the event]",
      "causal_link": "1"
    \},
    \{ ... \}
  ]
\}
\end{alltt}
\end{tcolorbox}

\end{tcolorbox}
\captionof{figure}{Prompt template for generating temporally coherent event timelines.}
\label{fig:prompt_event_generation}
\end{center}

\clearpage
\subsection{Conversation Generation}
\begin{center}
\begin{tcolorbox}[
    promptbox,
    title={\textbf{Generating target-person conversations.}}
]
\footnotesize

\textbf{Role:} You are the Visual Reasoning Dataset Generator.

\textbf{Task:} Construct a 10 to 12 turn chat log between a User and an Assistant based on the inputs below. The goal is to test an AI's ability to ground visual context by recognizing a specific recurring person from the user's social graph, or the user themselves, in a past special event photo.

\textbf{Inputs:}
\begin{itemize}[leftmargin=1em, itemsep=0pt, topsep=0pt, parsep=0pt]
    \item \textbf{User Persona:} \textcolor{red}{\{persona\_json\}}
    \item \textbf{Special Event:} \textcolor{red}{\{event\_summary\}}
    \item \textbf{Event Date:} \textcolor{red}{\{event\_date\}}
    \item \textbf{Target Person ID:} \textcolor{red}{\{target\_person\_id\}}
\end{itemize}

\textbf{Step-by-Step Generation Rules:}
\begin{enumerate}[leftmargin=1.2em, itemsep=0pt, topsep=0pt, parsep=0pt]
    \item \textbf{The Visual Prompt}
    \begin{itemize}[leftmargin=1em, itemsep=0pt, topsep=0pt, parsep=0pt]
        \item Generate a dense, hyper-realistic image prompt. It will describe the main focus and the environment based on the Special Event.
        \item \textbf{Target Injection:}
        \begin{itemize}[leftmargin=1em, itemsep=0pt, topsep=0pt, parsep=0pt]
            \item \textbf{If Target Person ID is \texttt{<main>}:} The image MUST be a \textbf{Third-Person or Selfie shot} focusing on the main user, \texttt{<main>}. The user must be fully visible in the scene, posing or participating in the event. Do NOT add a second person or a duplicate \texttt{<main>}.
            \item \textbf{Otherwise:} The image MUST be a \textbf{Third-Person or Selfie shot}. It MUST explicitly include both the main user, \texttt{<main>}, and the Target Person, \textcolor{red}{\{target\_person\_id\}}. Both characters must be fully visible in the scene, interacting, posing, or participating in the event together.
        \end{itemize}
        \item \textbf{Facial Features vs. Clothing:} Do NOT describe physical facial features. The system loads these automatically. You MUST describe clothing, pose, and actions in rich detail.
    \end{itemize}

    \item \textbf{The Conversation}
    \begin{itemize}[leftmargin=1em, itemsep=0pt, topsep=0pt, parsep=0pt]
        \item Write a natural chat between a User and an Assistant freely discussing the uploaded Special Event photo.
        \item \textbf{Always Asking:} The User is NOT telling a story; they are seeking help, reminiscing, asking for opinions, or checking details about the event. Almost every User turn should be a question.
        \item \textbf{Photo Context Rule:}
        \begin{itemize}[leftmargin=1em, itemsep=0pt, topsep=0pt, parsep=0pt]
            \item \textbf{If Target Person ID is \texttt{<main>}:} The User MUST establish that they are uploading a photo of themselves from that past event, e.g., ``Look at this picture of me at my birthday''.
            \item \textbf{Otherwise:} The User MUST establish that they are uploading a photo of themselves and the other person from that past event, e.g., ``Look at this picture of us from my birthday'' or ``Do you remember when we went here?''.
        \end{itemize}
        \item \textbf{Identity Constraint:}
        \begin{itemize}[leftmargin=1em, itemsep=0pt, topsep=0pt, parsep=0pt]
            \item \textbf{If Target Person ID is \texttt{<main>}:} The User and Assistant refer to the user in the first/second person, e.g., ``I'', ``me'', or ``you''. Do not refer to \texttt{<main>} as a third party.
            \item \textbf{Otherwise:} Neither the User nor the Assistant can explicitly state the Target Person's name or the user's relationship to them, e.g., never say ``my friend'', ``my coworker'', ``my partner'', or ``David''. Refer only using pronouns like ``he'', ``she'', ``they'', or ``we''.
        \end{itemize}
        \item \textbf{Implicit Familiarity:} The conversation should imply that the Assistant already knows the visual context.
    \end{itemize}
\end{enumerate}

\textbf{Formatting Rules:}
\begin{enumerate}[leftmargin=1.2em, itemsep=0pt, topsep=0pt, parsep=0pt]
    \item \textbf{Image Format:} \texttt{`<image> [Visual Prompt] </image> [User Text]'}
    \item \textbf{Strict Character Tags:} Inside the \texttt{<image>} tags, you MUST NOT use names. Substitute them exactly as provided, e.g., \texttt{<main>} or \textcolor{red}{\{target\_person\_id\}}.
\end{enumerate}

\textbf{Output Format:} Return ONLY valid JSON.

\vspace{0.2em}
\begin{tcolorbox}[promptoutputbox]
\scriptsize
\begin{alltt}
\{
  "question_hint": \{
    "target_person_id": "\textcolor{red}{\{target\_person\_id\}}",
    "identity_concealment_check": "[Briefly quote the text showing how
                                   identity was handled, e.g., 'Look at my outfit']"
  \},
  "conversation": [
    \{"role": "user", "content": "..."\},
    \{"role": "assistant", "content": "..."\},
  ]
\}
\end{alltt}
\end{tcolorbox}

\end{tcolorbox}
\captionof{figure}{Prompt template for generating target-person conversations.}
\label{fig:prompt_target_person_conversation_generation}
\end{center}

\clearpage
\begin{center}
\begin{tcolorbox}[
    promptbox,
    title={\textbf{Generating target-asset conversations.}}
]
\footnotesize

\textbf{Role:} You are the Visual Reasoning Dataset Generator.

\textbf{Task:} Construct a 10 to 12 turn chat log between a User and an Assistant based on the inputs below. The goal is to test an AI's ability to ``remember'' visual details from a past image that were never mentioned in the text.

\textbf{Inputs:}
\begin{itemize}[leftmargin=1em, itemsep=0pt, topsep=0pt, parsep=0pt]
    \item \textbf{User Persona:} \textcolor{red}{\{persona\_json\}}
    \item \textbf{Event:} \textcolor{red}{\{event\_summary\}}
    \item \textbf{Date:} \textcolor{red}{\{event\_date\}}
    \item \textbf{Target Asset:} \textcolor{red}{\{target\_asset\}}
\end{itemize}

\textbf{Step-by-Step Generation Rules:}
\begin{enumerate}[leftmargin=1.2em, itemsep=0pt, topsep=0pt, parsep=0pt]
    \item \textbf{The Visual Prompt}
    \begin{itemize}[leftmargin=1em, itemsep=0pt, topsep=0pt, parsep=0pt]
        \item Generate a dense, hyper-realistic image prompt. It will describe the main focus on the foreground and the incidental background details.
        \item \textbf{Target Injection \& Camera Perspective:} The camera perspective is strictly determined by how the Target Asset is physically interacted with:
        \begin{itemize}[leftmargin=1em, itemsep=0pt, topsep=0pt, parsep=0pt]
            \item \textbf{Condition A (Held or Worn $\rightarrow$ Third-Person):} If the Target Asset is an item the user holds or wears, the image MUST be a \textbf{Third-Person} shot. The \texttt{<main>} tag must be fully visible in the scene, and you must explicitly describe \texttt{<main>} wearing or holding the item to unambiguously establish visual ownership. \textbf{Crucial:} There MUST NOT be a second person, e.g., \texttt{<0\_0>}, in this image to avoid ambiguity about who the object belongs to.
            \item \textbf{Condition B (Environmental $\rightarrow$ Scene Shot):} If the Target Asset is placed in the environment, e.g., on a desk or floor, the image MUST be a standard scene shot. Do NOT use terms like ``POV'' or ``First-Person'' in the image prompt. Do NOT include the \texttt{<main>} tag or any reference to the main user. The asset must be immediately in their personal workspace/home/room.
        \end{itemize}
        \item \textbf{No Detailed Descriptions for the Target:} Do NOT describe the visual specifics of the Target Asset in the prompt. Refer to it using a simple, generic term, e.g., ``a watch'' or ``a laptop''. The exact visual details will be handled externally via image conditioning.
    \end{itemize}

    \item \textbf{The Conversation}
    \begin{itemize}[leftmargin=1em, itemsep=0pt, topsep=0pt, parsep=0pt]
        \item Write a natural chat between a User and an Assistant about the Event.
        \item \textbf{Always Asking:} The User is NOT telling a story; they are seeking help, opinions, or checking details. Almost every User turn should be a question.
        \item \textbf{Perspective Context Rule:} The text must align with the chosen camera perspective:
        \begin{itemize}[leftmargin=1em, itemsep=0pt, topsep=0pt, parsep=0pt]
            \item \textbf{If Third-Person (Condition A):} The User MUST explicitly mention they are sharing a past photo taken by someone else, e.g., ``Look at this picture my friend took of me''.
            \item \textbf{If Scene Shot (Condition B):} The User should naturally refer to the space, e.g., ``Here is my desk'' or ``This is my room''. Do NOT use phrases like ``Here is what I am looking at''.
        \end{itemize}
        \item \textbf{Ownership Constraint:}
        \begin{itemize}[leftmargin=1em, itemsep=0pt, topsep=0pt, parsep=0pt]
            \item \textbf{If Third-Person (Condition A):} You do NOT need to establish implicit ownership in the text, as it is already established visually. Just follow the Perspective Context Rule.
            \item \textbf{If Scene Shot (Condition B):} The User must provide conversational context that implicitly establishes their ownership of the environment, e.g., ``Here is a picture of my working table'' or ``This is my room''.
        \end{itemize}
        \item \textbf{Hard Constraint:} Neither the User nor the Assistant can explicitly name, list, or describe the Target Asset itself.
        \item The Visual Prompt can be in any turn, not necessarily the first turn.
    \end{itemize}
\end{enumerate}

\textbf{Formatting Rules:}
\begin{enumerate}[leftmargin=1.2em, itemsep=0pt, topsep=0pt, parsep=0pt]
    \item \textbf{Image Format:} \texttt{`<image> [Visual Prompt] </image> [User Text]'}
    \item \textbf{Strict Character Tags:} Inside the \texttt{<image>} tags, you MUST NOT use names. Substitute them as follows:
    \begin{itemize}[leftmargin=1em, itemsep=0pt, topsep=0pt, parsep=0pt]
        \item Refer to the Main User as \texttt{<main>}.
        \item Refer to social connections ONLY by their ID key, e.g., \texttt{<0\_0>} or \texttt{<0\_1>}.
    \end{itemize}
    \item \textbf{Visual Content \& Clothing:} You MUST describe the clothing/outfit for every character present in the tag, e.g., \texttt{<0\_1>} wearing a lab coat. Do not describe facial features.
\end{enumerate}

\textbf{Output Format:} Return ONLY valid JSON.

\vspace{0.2em}
\begin{tcolorbox}[promptoutputbox]
\scriptsize
\begin{alltt}
\{
  "question_hint": \{
    "target_asset_id": "[Extract the asset_id from the Target Asset]",
    "generic_reference_used": "[The simple term used in the prompt, e.g., 'a watch']",
    "perspective_used": "[State either 'Scene Shot' or 'Third-Person']",
    "ownership_context": "[Briefly quote how the user implied ownership in
                          text, or write 'N/A' if Condition A]"
  \},
  "conversation": [
    \{"role": "user", "content": "..."\},
    \{"role": "assistant", "content": "..."\},
  ]
\}
\end{alltt}
\end{tcolorbox}

\end{tcolorbox}
\captionof{figure}{Prompt template for generating target-asset conversations.}
\label{fig:prompt_target_asset_conversation_generation}
\end{center}

\clearpage
\begin{center}
\centering   
\begin{tcolorbox}[
    promptbox,
    title={\textbf{Generating implicit-fact (visual-only) conversations.}}
]
\footnotesize
\textbf{Role:} You are the Visual Forensic Scenario Architect.

\textbf{Task:} Construct a 10 to 12 turn chat log between a User and an Assistant based on the inputs below. The scenario is containing subtle ``Visual Proxies'' that prove an \textbf{Implicit Fact} about the user.

\textbf{Inputs:}
\begin{itemize}[leftmargin=1em, itemsep=0pt, topsep=0pt, parsep=0pt]
    \item \textbf{User Persona:} \textcolor{red}{\{persona\_json\}}
    \item \textbf{Event:} \textcolor{red}{\{event\_summary\}}
    \item \textbf{Date:} \textcolor{red}{\{event\_date\}}
    \item \textbf{ALREADY USED Implicit Facts (DO NOT REPEAT):} \textcolor{red}{\{used\_implicit\_facts\}}
\end{itemize}

\textbf{Step-by-Step Generation Rules:}
\begin{enumerate}[leftmargin=1.2em, itemsep=0pt, topsep=0pt, parsep=0pt]
    \item \textbf{Construct the Deduction Chain}
    \begin{itemize}[leftmargin=0.5em, itemsep=1pt, topsep=1pt]
        \item \textbf{Implicit Fact:} Invent a specific fact about the user, the other characters, or their situation.
        \item \textbf{CRITICAL (Diversity):} Analyze the categories of the ``ALREADY USED Implicit Facts'' list and select a fact from a completely different semantic domain.
        \item \textbf{CRITICAL (Uniqueness):} Do NOT use any facts similar to those listed in ``ALREADY USED''.
        \item \textbf{CRITICAL (Consistency):} The new fact must not logically contradict the ``ALREADY USED'' facts.
        \item \textbf{Visible Proxies:} Invent 1--2 specific visual details that logically prove this fact.
        \item \textbf{The Logic:} The deduction must be definitive; the Implicit Fact must be the only reasonable conclusion.
        \item \textbf{Anti-Hallucination:} The \texttt{implicit\_fact} must not contain assumptions or embellishments not proven by the proxies.
        \item \textbf{HARD BAN on Secrets:} Do NOT use ``secretly'' or frame this fact as an implicit secret, taboo, or dramatic revelation.
    \end{itemize}

    \item \textbf{The Visual Prompt}
    \begin{itemize}[leftmargin=1em, itemsep=0pt, topsep=0pt, parsep=0pt]
        \item Generate a dense, hyper-realistic image prompt describing the environment.
        \item Naturally integrate the Visible Proxies into the scene description; do not highlight them as clues.
        \item \textbf{Camera Perspective \& Subjects:} Explicitly define the shot type.
        \begin{itemize}[leftmargin=1em, itemsep=0pt, topsep=0pt, parsep=0pt]
            \item \textbf{POV Shot:} The user is behind the camera. Do NOT include the \texttt{<main>} tag.
            \item \textbf{Third-Person Shot:} The user is in the photo. Include the \texttt{<main>} tag and describe their actions/clothing.
            \item \textbf{THE PHOTOGRAPHER RULE:} The person holding the camera cannot be in the picture.
        \end{itemize}
    \end{itemize}

    \item \textbf{The Conversation}
    \begin{itemize}[leftmargin=1em, itemsep=0pt, topsep=0pt, parsep=0pt]
        \item Write a natural chat between the User and an Assistant about the Event.
        \item The Visual Prompt can be in any turn, not necessarily the first turn.
        \item \textbf{Natural Dialogue:} Outside the \texttt{<image>} block, use natural names or relationship terms; do NOT use tags like \texttt{<main>} or \texttt{<0\_0>}.
        \item \textbf{Perspective Match:} The User's text must logically match the camera perspective established in the Visual Prompt.
        \item \textbf{Always Asking:} The User is NOT telling a story; almost every User turn should be a question.
        \item \textbf{Constraint:} The User acts normally and does not explain their habits or the setup.
        \item \textbf{Hard Ban:} Do NOT mention the Implicit Fact directly.
    \end{itemize}
\end{enumerate}

\textbf{Formatting Rules:}
 \begin{enumerate}[leftmargin=1.2em, itemsep=0pt, topsep=0pt, parsep=0pt]
    \item \textbf{Image Format:} \texttt{`<image> [Visual Prompt] </image> [User Text]'}
    \item \textbf{Strict Tag Boundaries:}
    \begin{itemize}[leftmargin=1em, itemsep=0pt, topsep=0pt, parsep=0pt]
        \item \textbf{Inside \texttt{<image>} tags:} Use only \texttt{<main>}, \texttt{<0\_0>}, \texttt{<0\_1>}, etc.; do NOT use natural names.
        \item \textbf{Outside \texttt{<image>} tags:} Use natural names or relationship identifiers; do NOT use structural tags.
    \end{itemize}
    \item \textbf{Visual Content \& Clothing:}
    \begin{itemize}[leftmargin=1em, itemsep=0pt, topsep=0pt, parsep=0pt]
        \item Describe the clothing/outfit for every character present in the tag.
        \item Describe pose, action, and environment. Do NOT describe physical facial features.
    \end{itemize}
\end{enumerate}

\textbf{Output Format:} Return ONLY valid JSON.

\vspace{0.5em}
\begin{tcolorbox}[promptoutputbox]
\scriptsize
\begin{alltt}
\{
  "question_hint": \{
    "implicit_fact": "[The specific attribute, e.g., User is Left-Handed]",
    "visible_proxies": ["[Cue 1]", "[Cue 2]"],
    "world_knowledge_rule": "[The logic linking cues to fact]"
  \},
  "conversation": [
    \{"role": "user", "content": "..."\},
    \{"role": "assistant", "content": "..."\}
  ]
\}
\end{alltt}
\end{tcolorbox}

\end{tcolorbox}
\captionof{figure}{Prompt template for generating implicit-fact (visual-only) conversations.}
\label{fig:prompt_implicit_fact_vo_conversation_generation}
\end{center}

\clearpage
\begin{center}
\begin{tcolorbox}[
    promptbox,
    title={\textbf{Generating implicit-fact (multimodal) conversations.}}
]
\footnotesize

\textbf{Role:} You are the Multimodal Forensic Scenario Architect.

\textbf{Task:} Construct a 10 to 12 turn chat log between a User and an Assistant based on the inputs below. The scenario must contain an \textbf{Implicit Fact} about the user that can ONLY be deduced by combining 1 Visual Cue and 1 Textual Cue.

\textbf{Inputs:}
\begin{itemize}[leftmargin=1em, itemsep=0pt, topsep=0pt, parsep=0pt]
    \item \textbf{User Persona:} \textcolor{red}{\{persona\_json\}}
    \item \textbf{Event:} \textcolor{red}{\{event\_summary\}}
    \item \textbf{Date:} \textcolor{red}{\{event\_date\}}
    \item \textbf{ALREADY USED Implicit Facts (DO NOT REPEAT):} \textcolor{red}{\{used\_implicit\_facts\}}
\end{itemize}

\textbf{Step-by-Step Generation Rules:}
\begin{enumerate}[leftmargin=1.2em, itemsep=0pt, topsep=0pt, parsep=0pt]
    \item \textbf{Construct the Split-Evidence Chain (The ``Lock and Key'' Logic)}
    \begin{itemize}[leftmargin=1em, itemsep=0pt, topsep=0pt, parsep=0pt]
        \item \textbf{Implicit Fact:} Invent a specific fact about the user, the other characters, or their situation.
        \item \textbf{CRITICAL (Diversity):} Analyze the categories of the ``ALREADY USED Implicit Facts'' list and select a fact from a completely different semantic domain.
        \item \textbf{CRITICAL (Uniqueness):} Do NOT use any facts similar to those listed in ``ALREADY USED''.
        \item \textbf{CRITICAL (Consistency):} The new fact must not logically contradict the ``ALREADY USED'' facts.
        \item \textbf{The Rejection Filter:} The fact must share 0\% thematic, structural, or conceptual overlap with the ``ALREADY USED'' list.
        \item \textbf{The Split:} Prove this fact using two distinct pieces of evidence that act as a Lock and Key:
        \begin{itemize}[leftmargin=1em, itemsep=0pt, topsep=0pt, parsep=0pt]
            \item \textbf{The Visual Proxy (The Lock):} An environmental detail or object that is highly specific but lacks context.
            \item \textbf{The Textual Proxy (The Key):} A specific phrase that, when applied to the visual proxy, eliminates all other alternative explanations.
        \end{itemize}
        \item \textbf{The Logic:} The deduction must be definitive; the Implicit Fact must be the only reasonable conclusion.
        \item \textbf{HARD BAN on Secrets:} Do NOT use ``secretly'' or frame this fact as an implicit secret, taboo, or dramatic revelation.
    \end{itemize}

    \item \textbf{The Visual Prompt}
    \begin{itemize}[leftmargin=1em, itemsep=0pt, topsep=0pt, parsep=0pt]
        \item Generate a dense, hyper-realistic image prompt describing the environment.
        \item Naturally integrate the Visible Proxies into the scene description; do not highlight them as clues.
        \item \textbf{Camera Perspective \& Subjects:} Explicitly define the shot type.
        \begin{itemize}[leftmargin=1em, itemsep=0pt, topsep=0pt, parsep=0pt]
            \item \textbf{POV Shot:} The user is behind the camera. Do NOT include the \texttt{<main>} tag.
            \item \textbf{Third-Person Shot:} The user is in the photo. Include the \texttt{<main>} tag and describe their actions/clothing.
            \item \textbf{THE PHOTOGRAPHER RULE:} The person holding the camera cannot be in the picture.
        \end{itemize}
    \end{itemize}

    \item \textbf{The Conversation}
    \begin{itemize}[leftmargin=1em, itemsep=0pt, topsep=0pt, parsep=0pt]
        \item Write a natural chat between the User and an Assistant about the Event.
        \item \textbf{CRITICAL (Separation):} The Visual Proxy and Textual Proxy MUST occur in different turns.
        \begin{itemize}[leftmargin=1em, itemsep=0pt, topsep=0pt, parsep=0pt]
            \item Do not put the text clue in the caption of the image.
            \item Do not put the text clue in the message immediately following the image.
            \item Separate them by at least 1 turn if possible.
        \end{itemize}
        \item The Visual Prompt can be in any turn, not necessarily the first turn.
        \item \textbf{Natural Dialogue:} Outside the \texttt{<image>} block, use natural names or relationship terms; do NOT use tags like \texttt{<main>} or \texttt{<0\_0>}.
        \item \textbf{Perspective Match:} The User's text must logically match the camera perspective established in the Visual Prompt.
        \item \textbf{Always Asking:} The User is NOT telling a story; almost every User turn should be a question.
        \item \textbf{Constraint:} The User acts normally and does not explain their habits or the setup.
        \item \textbf{Hard Ban:} Do NOT mention the Implicit Fact directly.
    \end{itemize}
\end{enumerate}

\textbf{Formatting Rules:}
\begin{enumerate}[leftmargin=1.2em, itemsep=0pt, topsep=0pt, parsep=0pt]
    \item \textbf{Image Format:} \texttt{`<image> [Visual Prompt] </image> [User Text]'}
    \item \textbf{Strict Tag Boundaries:}
    \begin{itemize}[leftmargin=1em, itemsep=0pt, topsep=0pt, parsep=0pt]
        \item \textbf{Inside \texttt{<image>} tags:} Use only \texttt{<main>}, \texttt{<0\_0>}, \texttt{<0\_1>}, etc.; do NOT use natural names.
        \item \textbf{Outside \texttt{<image>} tags:} Use natural names or relationship identifiers; do NOT use structural tags.
    \end{itemize}
    \item \textbf{Visual Content \& Clothing:} Describe clothing/outfit, pose, action, and environment for every tagged character. Do NOT describe physical facial features.
\end{enumerate}

\textbf{Output Format:} Return ONLY valid JSON.

\vspace{0.2em}
\begin{tcolorbox}[promptoutputbox]
\scriptsize
\begin{alltt}
\{
  "question_hint": \{
    "implicit_fact": "[The specific attribute, e.g., User plays Soccer]",
    "visual_proxy": "[The visual object, e.g., Muddy Cleats]",
    "textual_proxy": "[The quote, e.g., 'Sore from practice']",
    "world_knowledge_rule": "[The logic linking the two inputs to the fact]"
  \},
  "conversation": [
    \{"role": "user", "content": "..."\},
    \{"role": "assistant", "content": "..."\}
  ]
\}
\end{alltt}
\end{tcolorbox}

\end{tcolorbox}
\captionof{figure}{Prompt template for generating implicit-fact (multimodal) conversations.}
\label{fig:prompt_conversation_generation_hard_multimodal}
\end{center}

\clearpage
\begin{center}
\begin{tcolorbox}[
    promptbox,
    title={\textbf{Generating neutral conversations.}}
]
\footnotesize

\textbf{Role:} You are the Visual Reasoning Dataset Generator.

\textbf{Task:} Construct a 10 to 12 turn chat log between a User and an Assistant based on the inputs below. This is a ``neutral'' baseline conversation. The goal is simply to create a natural, engaging conversation grounded in the provided event and a corresponding image.

\textbf{Inputs:}
\begin{itemize}[leftmargin=1em, itemsep=0pt, topsep=0pt, parsep=0pt]
    \item \textbf{User Persona:} \textcolor{red}{\{persona\_json\}}
    \item \textbf{Event:} \textcolor{red}{\{event\_summary\}}
    \item \textbf{Date:} \textcolor{red}{\{event\_date\}}
\end{itemize}

\textbf{Step-by-Step Generation Rules:}
\begin{enumerate}[leftmargin=1.2em, itemsep=0pt, topsep=0pt, parsep=0pt]
    \item \textbf{The Visual Prompt}
    \begin{itemize}[leftmargin=1em, itemsep=0pt, topsep=0pt, parsep=0pt]
        \item Generate a dense, hyper-realistic image prompt. It will describe the main focus and the environment based naturally on the Event.
        \item \textbf{Location Constraint:} The setting MUST NOT be the main user's own home or personal private space. The image and event must explicitly take place outside, in a public space, e.g., a cafe, park, store, or venue, a professional setting, or at someone else's home.
        \item \textbf{Camera Perspective:} Mostly, the image should be a \textbf{Scene Shot}. Do NOT use terms like ``POV'' or ``First-Person'' in the image prompt. Do NOT include the \texttt{<main>} tag or any reference to the main user in the prompt.
        \item \textbf{Alternative Perspective:} If the event naturally requires the user to be seen, e.g., showing off an outfit or performing a full-body action, you may generate a \textbf{Third-Person} shot where \texttt{<main>} is fully visible in the scene.
    \end{itemize}

    \item \textbf{The Conversation}
    \begin{itemize}[leftmargin=1em, itemsep=0pt, topsep=0pt, parsep=0pt]
        \item Write a natural chat between a User and an Assistant about the Event.
        \item \textbf{Always Asking:} The User is NOT telling a story; they are seeking help, opinions, or checking details. Almost every User turn should be a question.
        \item \textbf{Perspective Context Rule:} The text must align with the chosen camera perspective:
        \begin{itemize}[leftmargin=1em, itemsep=0pt, topsep=0pt, parsep=0pt]
            \item \textbf{If Third-Person:} The User MUST explicitly mention they are sharing a past photo taken by someone else, e.g., ``Look at this picture my friend took of me''.
            \item \textbf{If Scene Shot:} The User should introduce the image casually, as if texting a photo to a friend. Use natural, conversational phrasing or simply start discussing the activity shown. Do NOT use stiff announcements like ``Here is the venue,'' ``Look at this setup,'' or ``Here is a picture of what I am looking at.''
        \end{itemize}
        \item \textbf{Constraint:} The User focuses only on the main activity, emotion, or event shown in the image or described in the summary.
        \item The Visual Prompt can be in any turn, not necessarily the first turn.
    \end{itemize}
\end{enumerate}

\textbf{Formatting Rules:}
\begin{enumerate}[leftmargin=1.2em, itemsep=0pt, topsep=0pt, parsep=0pt]
    \item \textbf{Image Format:} \texttt{`<image> [Visual Prompt] </image> [User Text]'}
    \item \textbf{Strict Character Tags:} Inside the \texttt{<image>} tags, you MUST NOT use names. Substitute them as follows:
    \begin{itemize}[leftmargin=1em, itemsep=0pt, topsep=0pt, parsep=0pt]
        \item Refer to the Main User as \texttt{<main>}.
        \item Refer to social connections ONLY by their ID key, e.g., \texttt{<0\_0>} or \texttt{<0\_1>}.
    \end{itemize}
    \item \textbf{Visual Content \& Clothing:} You MUST describe the clothing/outfit for every character present in the tag, e.g., \texttt{<0\_1>} wearing a lab coat. Do not describe physical facial features.
\end{enumerate}

\textbf{Output Format:} Return ONLY valid JSON.

\vspace{0.2em}
\begin{tcolorbox}[promptoutputbox]
\scriptsize
\begin{alltt}
\{
  "conversation": [
    \{"role": "user", "content": "..."\},
    \{"role": "assistant", "content": "..."\},
  ]
\}
\end{alltt}
\end{tcolorbox}

\end{tcolorbox}
\captionof{figure}{Prompt template for generating neutral conversations.}
\label{fig:prompt_neutral_conversation_generation}
\end{center}

\clearpage
\begin{center}
\begin{tcolorbox}[
    promptbox,
    title={\textbf{Generating hard-negative asset conversations.}}
]
\footnotesize

\textbf{Role:} You are the Visual Reasoning Dataset Generator.

\textbf{Task:} Construct a 10 to 12 turn chat log between a User and an Assistant based on the inputs below. The goal is to test an AI's ability to ground visual context when an object explicitly belongs to someone else or is in a completely different environment.

\textbf{Inputs:}
\begin{itemize}[leftmargin=1em, itemsep=0pt, topsep=0pt, parsep=0pt]
    \item \textbf{User Persona:} \textcolor{red}{\{persona\_json\}}
    \item \textbf{Event:} \textcolor{red}{\{event\_summary\}}
    \item \textbf{Date:} \textcolor{red}{\{event\_date\}}
    \item \textbf{Variant Asset:} \textcolor{red}{\{variant\_asset\}}
\end{itemize}

\textbf{Step-by-Step Generation Rules:}
\begin{enumerate}[leftmargin=1.2em, itemsep=0pt, topsep=0pt, parsep=0pt]
    \item \textbf{The Visual Prompt}
    \begin{itemize}[leftmargin=1em, itemsep=0pt, topsep=0pt, parsep=0pt]
        \item Generate a dense, hyper-realistic image prompt describing the main focus and background.
        \item \textbf{Camera Perspective:} By default, the image MUST be a \textbf{Scene Shot} from the user's perspective. Do NOT use terms like ``POV'' or ``First-Person'' in the image prompt. Do NOT include the \texttt{<main>} tag or any reference to the main user.
        \item \textbf{Alternative Perspective:} You may occasionally generate a \textbf{Third-Person} shot where \texttt{<main>} is fully visible in the scene, but you must follow the specific Conversation constraint below.
        \item \textbf{Variant Injection \& Visual Disassociation:} Include the Variant Asset in the scene. To prevent ambiguity, its physical placement MUST unambiguously show it does NOT belong to the user.
        \begin{itemize}[leftmargin=1em, itemsep=0pt, topsep=0pt, parsep=0pt]
            \item \textbf{If worn/held:} It must be worn or held by someone else, e.g., \texttt{<0\_0>}.
            \item \textbf{If environmental:} It must be placed in a way that implies someone else's ownership or a public space, e.g., on a coworker's desk or on a store shelf.
        \end{itemize}
        \item \textbf{No Detailed Descriptions for the Variant:} Do NOT describe the visual specifics of the Variant Asset in the prompt. Refer to it using a simple, generic term, e.g., ``a cat'', ``a laptop'', or ``a backpack''. The exact visual details will be handled externally.
    \end{itemize}

    \item \textbf{The Conversation}
    \begin{itemize}[leftmargin=1em, itemsep=0pt, topsep=0pt, parsep=0pt]
        \item Write a natural chat between a User and an Assistant about the Event.
        \item \textbf{Always Asking:} The User is NOT telling a story; they are seeking help, opinions, or checking details. Almost every User turn should be a question.
        \item \textbf{Perspective Context Rule:} The text must align with the chosen camera perspective:
        \begin{itemize}[leftmargin=1em, itemsep=0pt, topsep=0pt, parsep=0pt]
            \item \textbf{If Third-Person:} The User MUST explicitly mention they are sharing a past photo taken by someone else, e.g., ``Look at this picture my friend took of me''.
            \item \textbf{If Scene Shot:} The User should naturally refer to the space or the other person, e.g., ``Here is my coworker's desk'' or ``Look at this cat''. Do NOT use phrases like ``Here is what I am looking at''.
        \end{itemize}
        \item \textbf{Non-Ownership Constraint:} The User must provide conversational context that explicitly or implicitly establishes the Variant Asset or the environment belongs to \textbf{someone else} or is a public space, e.g., ``I was at my coworker's desk'', ``Look at this cat I saw at the shelter'', or ``My friend's setup''. The User MUST NOT claim ownership of the asset or the space.
        \item \textbf{Hard Constraint:} Neither the User nor the Assistant can explicitly describe the Variant Asset's specific visual details. The text must establish the other person's ownership, but leave the visual specifics to the image.
        \item The Visual Prompt can be in any turn, not necessarily the first turn.
    \end{itemize}
\end{enumerate}

\textbf{Formatting Rules:}
\begin{enumerate}[leftmargin=1.2em, itemsep=0pt, topsep=0pt, parsep=0pt]
    \item \textbf{Image Format:} \texttt{`<image> [Visual Prompt] </image> [User Text]'}
    \item \textbf{Strict Character Tags:} Inside the \texttt{<image>} tags, you MUST NOT use names. Substitute them as follows:
    \begin{itemize}[leftmargin=1em, itemsep=0pt, topsep=0pt, parsep=0pt]
        \item Refer to the Main User as \texttt{<main>}.
        \item Refer to social connections ONLY by their ID key, e.g., \texttt{<0\_0>} or \texttt{<0\_1>}.
    \end{itemize}
    \item \textbf{Visual Content \& Clothing:} You MUST describe the clothing/outfit for every character present in the tag, e.g., \texttt{<0\_1>} wearing a lab coat. Do not describe physical facial features.
\end{enumerate}

\textbf{Output Format:} Return ONLY valid JSON.

\vspace{0.2em}
\begin{tcolorbox}[promptoutputbox]
\scriptsize
\begin{alltt}
\{
  "variant_hint": \{
    "variant_asset_id": "[Extract the variation_id from the Variant Asset]",
    "generic_reference_used": "[The simple term you used in the prompt,
                               e.g., 'a laptop']",
    "perspective_used": "[State either 'Scene Shot' or 'Third-Person']",
    "non_ownership_context": "[Briefly quote how the user implied it belongs
                              to someone else, e.g., 'At my coworker's desk']"
  \},
  "conversation": [
    \{"role": "user", "content": "..."\},
    \{"role": "assistant", "content": "..."\},
  ]
\}
\end{alltt}
\end{tcolorbox}

\end{tcolorbox}
\captionof{figure}{Prompt template for generating hard-negative asset conversations.}
\label{fig:prompt_variant_asset_conversation_generation}
\end{center}

\clearpage
\subsection{Question Generation}
\begin{center}
\begin{tcolorbox}[
    promptbox,
    title={\textbf{Generating target-person benchmark questions.}}
]
\footnotesize

\textbf{Role:} You are the Visual Memory Evaluation Generator for an AI benchmark.

\textbf{Task:} Based on the provided User Persona and Event Details, create a rigorous Multiple Choice Question (MCQ) to test an AI's visual memory. The AI being tested will NOT be shown the image again; it must rely entirely on its long-term memory of a past conversation.

\textbf{Inputs:}
\begin{itemize}[leftmargin=1em, itemsep=0pt, topsep=0pt, parsep=0pt]
    \item \textbf{User Persona (Social Graph):} \textcolor{red}{\{persona\_json\}}
    \item \textbf{Event Details:} \textcolor{red}{\{event\_json\}}
    \item \textbf{Image in that event:} \texttt{<image>}
\end{itemize}

\textbf{Step-by-Step Generation Rules:}
\begin{enumerate}[leftmargin=1.2em, itemsep=0pt, topsep=0pt, parsep=0pt]
    \item \textbf{Question Design}
    \begin{itemize}[leftmargin=1em, itemsep=0pt, topsep=0pt, parsep=0pt]
        \item \textbf{Branching Goal:}
        \begin{itemize}[leftmargin=1em, itemsep=0pt, topsep=0pt, parsep=0pt]
            \item \textbf{If Target Person ID is not \texttt{<main>}:} Test if the AI remembers the specific person from the user's social graph based on a past event image it saw previously.
            \item \textbf{If Target Person ID is \texttt{<main>}:} Test if the AI remembers specific visual details about the main user from a past event image it saw previously.
        \end{itemize}
        \item \textbf{Memory Framing:} Because the AI will not see the image during this test, frame the question as a recall task referencing a past interaction, e.g., ``In the photo I showed you earlier...'' or ``Do you remember the picture from...?''. Do NOT use phrases like ``in this photo'' or ``look at this image''.
        \item \textbf{Contextual Framing:} Include the specific event context in the question text using the provided Event Details, e.g., ``During my birthday party...''.
        \item \textbf{Question Text:}
        \begin{itemize}[leftmargin=1em, itemsep=0pt, topsep=0pt, parsep=0pt]
            \item \textbf{If Target Person ID is not \texttt{<main>}:} Ask a natural question about the past photo, e.g., ``Thinking back to the photo I showed you from my Thanksgiving dinner last year, who was the person standing with me?''.
            \item \textbf{If Target Person ID is \texttt{<main>}:} Ask a natural question about the user's appearance in the past event, e.g., ``Do you remember at my Thanksgiving dinner, what was I wearing?''.
        \end{itemize}
        \item Do NOT include the correct answer in the question text.
    \end{itemize}

    \item \textbf{Choices Design (Text Options)}
    \begin{itemize}[leftmargin=1em, itemsep=0pt, topsep=0pt, parsep=0pt]
        \item \textbf{Correct Answer:}
        \begin{itemize}[leftmargin=1em, itemsep=0pt, topsep=0pt, parsep=0pt]
            \item \textbf{If Target Person ID is not \texttt{<main>}:} The exact name of the person corresponding to the target ID in the User Persona, e.g., ``Liam''.
            \item \textbf{If Target Person ID is \texttt{<main>}:} A realistic text description that correctly identifies the user's clothing or action in that specific event.
        \end{itemize}
        \item \textbf{Distractors:}
        \begin{itemize}[leftmargin=1em, itemsep=0pt, topsep=0pt, parsep=0pt]
            \item \textbf{If Target Person ID is not \texttt{<main>}:} Names of other people from the user's social graph in the Persona, or highly plausible realistic fake relationships.
            \item \textbf{If Target Person ID is \texttt{<main>}:} Plausible but incorrect clothing, accessory, or action descriptions.
        \end{itemize}
        \item \textbf{Randomization:} Randomly assign the correct text to option A, B, C, or D.
    \end{itemize}
\end{enumerate}

\textbf{Output Format:} Return ONLY valid JSON.

\vspace{0.2em}
\begin{tcolorbox}[promptoutputbox]
\scriptsize
\begin{alltt}
\{
  "evaluation_mcq": [
    \{
      "id": 1,
      "question_type": "text",
      "question": "[Your generated memory recall question containing the event context]",
      "choices": \{
        "A": "[Text option]",
        "B": "[Text option]",
        "C": "[Text option]",
        "D": "[Text option]"
      \},
      "ground_truth": "[A, B, C, or D]"
    \}
  ]
\}
\end{alltt}
\end{tcolorbox}

\end{tcolorbox}
\captionof{figure}{Prompt template for generating target-person benchmark questions.}
\label{fig:prompt_target_person_mcq_generation}
\end{center}

\clearpage
\begin{center}
\begin{tcolorbox}[
    promptbox,
    title={\textbf{Generating target-asset benchmark questions.}}
]
\footnotesize

\textbf{Role:} You are the Visual Memory Evaluation Generator for an AI benchmark.

\textbf{Task:} Based on the provided Target Asset and its Variations, generate rigorous Multiple Choice Questions (MCQs) to test if an AI remembers the specific visual details of the user's object from a past conversation.

\textbf{Input:}
\begin{itemize}[leftmargin=1em, itemsep=0pt, topsep=0pt, parsep=0pt]
    \item \textbf{Asset Data:} \textcolor{red}{\{asset\_with\_variations\_json\}}
\end{itemize}

\textbf{Step-by-Step Generation Rules:}
\begin{enumerate}[leftmargin=1.2em, itemsep=0pt, topsep=0pt, parsep=0pt]
    \item \textbf{Question 1: Visual Recognition (Image Choices)}
    \begin{itemize}[leftmargin=1em, itemsep=0pt, topsep=0pt, parsep=0pt]
        \item \textbf{Goal:} Test if the AI can identify the exact object it saw previously among visually similar distractors.
        \item \textbf{Question Text:} Write a natural, concise question asking the AI to identify the user's item, e.g., ``Which of these travel tumblers is mine?'' or ``Can you point out my laptop?''. Do NOT include visual descriptors in the question text.
        \item \textbf{Choices:} The four options, A--D, must be the exact \texttt{image\_path} strings from the input data. One is the target asset's path, and the other three are the variation paths.
        \item \textbf{Randomization:} Randomly assign the correct \texttt{image\_path} to A, B, C, or D so it is not always the first option.
    \end{itemize}

    \item \textbf{Question 2: Detail Recall (Text Choices)}
    \begin{itemize}[leftmargin=1em, itemsep=0pt, topsep=0pt, parsep=0pt]
        \item \textbf{Goal:} Test if the AI remembers a specific, granular visual detail about the object without looking at the image again.
        \item \textbf{Question Text:} Ask about a highly specific feature of the target asset, e.g., ``What specific stickers are on my tumbler?'', ``What color is the lid of my tumbler?'', or ``Where is the scratch on my item?''.
        \item \textbf{Choices:} Write four distinct text options, A--D.
        \begin{itemize}[leftmargin=1em, itemsep=0pt, topsep=0pt, parsep=0pt]
            \item The correct option must accurately describe the detail based on the target asset's \texttt{visual\_description}.
            \item The three incorrect options, i.e., hard negatives, MUST be derived directly from the \texttt{visual\_description} of the three Variations.
        \end{itemize}
        \item \textbf{Randomization:} Randomly assign the correct text description to A, B, C, or D.
    \end{itemize}
\end{enumerate}

\textbf{Output Format:} Return ONLY valid JSON.

\vspace{0.2em}
\begin{tcolorbox}[promptoutputbox]
\scriptsize
\begin{alltt}
\{
  "evaluations": [
    \{
      "id": 1,
      "type": "image",
      "question": "[The visual recognition question]",
      "choices": \{
        "A": "[image_path]",
        "B": "[image_path]",
        "C": "[image_path]",
        "D": "[image_path]"
      \},
      "ground_truth": "[A, B, C, or D]"
    \},
    \{
      "id": 2,
      "type": "text",
      "question": "[The detail recall question]",
      "choices": \{
        "A": "[text description]",
        "B": "[text description]",
        "C": "[text description]",
        "D": "[text description]"
      \},
      "ground_truth": "[A, B, C, or D]"
    \}
    ... more question
  ]
\}
\end{alltt}
\end{tcolorbox}

\end{tcolorbox}
\captionof{figure}{Prompt template for generating target-asset benchmark questions.}
\label{fig:prompt_asset_visual_mcq_generation}
\end{center}

\clearpage
\begin{center}
\begin{tcolorbox}[
    promptbox,
    title={\textbf{Generating implicit-fact benchmark questions.}}
]
\footnotesize

\textbf{Role:} You are the Adversarial Benchmark Engineer. Your job is to create brutally difficult memory evaluations that easily fool standard, memoryless LLMs.

\textbf{Task:} Create 1 ``Future Questions'' in a Multiple Choice format with 4 options. The User returns days or months later with a \textbf{new problem}. To solve this problem, the Assistant MUST rely entirely on the \textbf{Implicit Fact}. If the AI does not know the fact, it MUST fail.

\textbf{Inputs:}
\begin{enumerate}[leftmargin=1.2em, itemsep=0pt, topsep=0pt, parsep=0pt]
    \item \textbf{The Implicit Fact:} \textcolor{red}{\{implicit\_fact\}}
    \item \textbf{User Persona:} \textcolor{red}{\{persona\_json\}}
    \item \textbf{Original Conversation Context:} \textcolor{red}{\{conversation\}}
\end{enumerate}

\textbf{Critical Constraints:}
\begin{enumerate}[leftmargin=1.2em, itemsep=0pt, topsep=0pt, parsep=0pt]
    \item \textbf{The ``Information Void'' Rule:}
    \begin{itemize}[leftmargin=1em, itemsep=0pt, topsep=0pt, parsep=0pt]
        \item The User asks a question with max 25 words.
        \item \textbf{NO LEXICAL LEAKS:} The question must NEVER use words that match the correct option.
        \item \textbf{NO CONTEXT CLUES:} Do not name the specific item the implicit fact relies on. Do not use phrases like ``Given my diet,'' ``For my condition,'' or ``As we discussed.''
    \end{itemize}

    \item \textbf{Weaponizing RLHF Bias}
    \begin{itemize}[leftmargin=1em, itemsep=0pt, topsep=0pt, parsep=0pt]
        \item \textbf{The ``RLHF Bait'' Distractors:} The 3 distractors must be the exact answers a helpful, harmless, and honest instruction-tuned LLM would want to give: best-practice, popular, healthy, polite, or logical advice for a generic user.
        \item \textbf{The ``Suboptimal Truth'' Correct Option:} To a generic AI, the correct option MUST look like the worst choice: counter-intuitive, boring, inconvenient, overly restrictive, or slightly unhelpful. Because of the Implicit Fact, it is the only valid choice.
        \item \textbf{Example:} If the Implicit Fact is ``User is deathly allergic to all vegetables,'' then a generic healthy dinner recommendation should prefer balanced options, while the correct option may be ``Just a plain steak with nothing else on the plate.''
    \end{itemize}

    \item \textbf{Equal Specificity:} All four options MUST have the same length, tone, and level of detail. If the correct answer is a specific 15-word description, the traps must also be specific 15-word descriptions.

    \item \textbf{No Context Matching:} The correct option must not share synonyms or environmental clues with the question.

    \item \textbf{Strict Randomization:} Randomize which letter, A--D, contains the correct answer. The ground truth should be equally likely to be A, B, C, or D.

    \item \textbf{Determine Query Modality:}
    \begin{itemize}[leftmargin=1em, itemsep=0pt, topsep=0pt, parsep=0pt]
        \item \textbf{Text Only:} User asks a question directly.
        \item \textbf{Image Based:} User uploads a new image and asks a question about it.
    \end{itemize}
\end{enumerate}

\textbf{Formatting Rules for Query Image Prompt}
\begin{enumerate}[leftmargin=1.2em, itemsep=0pt, topsep=0pt, parsep=0pt]
    \item \textbf{Format:} \texttt{`<image> [Visual Prompt] </image>'}
    \item \textbf{Tags:} Use \texttt{<main>} or IDs, e.g., \texttt{<0\_0>}, ONLY if the person is physically in the frame. If it is an object close-up, do not use tags.
    \item \textbf{Content:} Describe clothing/environment if tags are used.
\end{enumerate}

\textbf{Output Format:} Return ONLY valid JSON.

\vspace{0.2em}
\begin{tcolorbox}[promptoutputbox]
\scriptsize
\begin{alltt}
\{
  "questions": [
    \{
      "id": 1,
      "type": "text",
      "query_image_prompt": null,
      "question": "[The natural language question without any tags]",
      "choices": \{
        "A": "[Option text]",
        "B": "[Option text]",
        "C": "[Option text]",
        "D": "[Option text]"
      \},
      "ground_truth": "[A, B, C or D]",
      "reasoning_chain": "Generic AI trap: An LLM without memory will choose
                          [Trap] because it is [Healthy/Polite/Standard Best
                          Practice]. It will avoid [Correct Option] because
                          it looks [Boring/Suboptimal/Weird]. However, applying
                          the Implicit Fact, [Correct Option] is strictly required
                          because [Reason]."
    \}
  ]
\}
\end{alltt}
\end{tcolorbox}

\end{tcolorbox}
\captionof{figure}{Prompt template for generating implicit-fact benchmark questions.}
\label{fig:prompt_question_generation_hard_mcq}
\end{center}

\clearpage

\newpage
\section*{NeurIPS Paper Checklist}
\begin{enumerate}

\item {\bf Claims}
    \item[] Question: Do the main claims made in the abstract and introduction accurately reflect the paper's contributions and scope?
    \item[] Answer: \answerYes{}
    \item[] Justification: The abstract and introduction state the two main contributions of the paper: a visual memory benchmark and the \textsc{VisualMem} framework, and these claims are supported by the benchmark construction, method, and experimental results described in the paper (Abstract; Secs.~1, 3, 4, and 5).

\item {\bf Limitations}
    \item[] Question: Does the paper discuss the limitations of the work performed by the authors?
    \item[] Answer: \answerYes{}
    \item[] Justification: A dedicated Limitations section following the conclusion discusses synthetic data assumptions, model dependence, the limited number of personas and benchmarks, and remaining gaps to real-world multimodal interactions.

\item {\bf Theory assumptions and proofs}
    \item[] Question: For each theoretical result, does the paper provide the full set of assumptions and a complete (and correct) proof?
    \item[] Answer: \answerNA{}
    \item[] Justification: The paper does not present theoretical results, theorems, or proofs.

    \item {\bf Experimental result reproducibility}
    \item[] Question: Does the paper fully disclose all the information needed to reproduce the main experimental results of the paper to the extent that it affects the main claims and/or conclusions of the paper (regardless of whether the code and data are provided or not)?
    \item[] Answer: \answerYes{}
    \item[] Justification: The paper describes the benchmark construction pipeline, the \textsc{VisualMem} architecture, the evaluation settings, baselines, backbone models, and metrics used for the main results (Secs.~3, 4, and 5). Appendix~\ref{app:implementation} further specifies the implementation, storage, pending-resolution, and retrieval configurations.

\item {\bf Open access to data and code}
    \item[] Question: Does the paper provide open access to the data and code, with sufficient instructions to faithfully reproduce the main experimental results, as described in supplemental material?
    \item[] Answer: \answerYes{}
    \item[] Justification: The implementation and evaluation scripts are publicly available at \url{https://github.com/viettmab/VisualMem}, and the benchmark data are available at \url{https://huggingface.co/datasets/viettmab/visualmem}. The repository provides environment setup, data download, and evaluation instructions.

\item {\bf Experimental setting/details}
    \item[] Question: Does the paper specify all the training and test details (e.g., data splits, hyperparameters, how they were chosen, type of optimizer) necessary to understand the results?
    \item[] Answer: \answerYes{}
    \item[] Justification: The paper specifies the benchmark setup, evaluation settings, baselines, backbone models, metrics, and ablations needed to interpret the results (Secs.~4 and 5), with full implementation and retrieval details in Appendix~\ref{app:implementation}.

\item {\bf Experiment statistical significance}
    \item[] Question: Does the paper report error bars suitably and correctly defined or other appropriate information about the statistical significance of the experiments?
    \item[] Answer: \answerYes{}
    \item[] Justification: Appendix~D reports mean accuracy and standard deviation over five runs for three representative methods: Mem0, MemOS, and \textsc{VisualMem}. Each run repeats memory construction, retrieval, and answer generation.

\item {\bf Experiments compute resources}
    \item[] Question: For each experiment, does the paper provide sufficient information on the computer resources (type of compute workers, memory, time of execution) needed to reproduce the experiments?
    \item[] Answer: \answerNo{}
    \item[] Justification: The paper currently reports the models and evaluation protocol, but does not yet provide a detailed accounting of compute resources, runtime, or total cost.
    
\item {\bf Code of ethics}
    \item[] Question: Does the research conducted in the paper conform, in every respect, with the NeurIPS Code of Ethics \url{https://neurips.cc/public/EthicsGuidelines}?
    \item[] Answer: \answerYes{}
    \item[] Justification: To the best of our knowledge, the research conforms to the NeurIPS Code of Ethics. The work uses synthetic benchmark construction rather than real personal conversations, which helps reduce privacy risks.

\item {\bf Broader impacts}
    \item[] Question: Does the paper discuss both potential positive societal impacts and negative societal impacts of the work performed?
    \item[] Answer: \answerYes{}
    \item[] Justification: The broader impacts discussion is provided in the appendix, including both potential benefits and risks of improved visual memory.
    
\item {\bf Safeguards}
    \item[] Question: Does the paper describe safeguards that have been put in place for responsible release of data or models that have a high risk for misuse (e.g., pre-trained language models, image generators, or scraped datasets)?
    \item[] Answer: \answerYes{}
    \item[] Justification: The benchmark is constructed synthetically rather than from scraped personal user data, and generated images are manually reviewed for consistency and quality before inclusion (Sec.~3). This reduces privacy exposure and helps avoid releasing unsafe or low-quality artifacts.

\item {\bf Licenses for existing assets}
    \item[] Question: Are the creators or original owners of assets (e.g., code, data, models), used in the paper, properly credited and are the license and terms of use explicitly mentioned and properly respected?
    \item[] Answer: \answerYes{}
    \item[] Justification: Yes, we have properly cited the code, data and models used.

\item {\bf New assets}
    \item[] Question: Are new assets introduced in the paper well documented and is the documentation provided alongside the assets?
    \item[] Answer: \answerYes{}
    \item[] Justification: The paper introduces and publicly releases the VisualMem benchmark and its implementation. Benchmark construction and validation are documented in Sec.~3 and Appendix~B, while setup, data download, and evaluation instructions are provided at \url{https://github.com/viettmab/VisualMem}. The benchmark is hosted at \url{https://huggingface.co/datasets/viettmab/visualmem}.

\item {\bf Crowdsourcing and research with human subjects}
    \item[] Question: For crowdsourcing experiments and research with human subjects, does the paper include the full text of instructions given to participants and screenshots, if applicable, as well as details about compensation (if any)? 
    \item[] Answer: \answerNA{}
    \item[] Justification: The benchmark quality-control review was performed by three internal researchers. No external crowd workers or study participants were recruited, and the work does not involve research with human subjects.

\item {\bf Institutional review board (IRB) approvals or equivalent for research with human subjects}
    \item[] Question: Does the paper describe potential risks incurred by study participants, whether such risks were disclosed to the subjects, and whether Institutional Review Board (IRB) approvals (or an equivalent approval/review based on the requirements of your country or institution) were obtained?
    \item[] Answer: \answerNA{}
    \item[] Justification: The benchmark quality-control review was conducted by internal researchers and did not recruit external participants or collect human-subject data; therefore, IRB approval was not applicable.

\item {\bf Declaration of LLM usage}
    \item[] Question: Does the paper describe the usage of LLMs if it is an important, original, or non-standard component of the core methods in this research? Note that if the LLM is used only for writing, editing, or formatting purposes and does \emph{not} impact the core methodology, scientific rigor, or originality of the research, declaration is not required.
    \item[] Answer: \answerYes{}
    \item[] Justification: LLMs and VLMs are part of both the benchmark construction pipeline and the \textsc{VisualMem} framework, including conversation generation, image-grounded interpretation, and question answering, and this usage is described in the method and benchmark sections (Secs.~3, 4, and 5).

\end{enumerate}

\end{document}